\documentclass{article}
\pdfoutput=1

\usepackage{microtype}
\usepackage{graphicx}
\usepackage{subcaption}
\usepackage{booktabs}
\usepackage{hyperref}
\usepackage{algorithm}
\usepackage{algorithmic}
\usepackage{amsmath}
\usepackage{amssymb}
\usepackage{mathtools}
\usepackage{amsthm}
\usepackage{wrapfig}
\usepackage{multirow}
\usepackage{bbm}

\PassOptionsToPackage{numbers, compress}{natbib}

\usepackage[preprint]{corl_2026} 

\title{Latent Memory Palace: Reasoning for Control as Autoregressive Variational Inference}

\author{
  Chuning Zhu \\
  University of Washington
  \And
  Eva Xu \\
  University of Washington  \\
  \And
  Jose Barreiros \\
  Toyota Research Institute
  \AND
  Krishnan Srinivasan \\
  Toyota Research Institute \\
  \And
  Paarth Shah \\
  Toyota Research Institute \\
  \And
  Abhishek Gupta \\
  University of Washington
}

\begin{document}

\newcommand{\boldme}[1]{{\footnotesize\textbf{#1}}}
\newcommand{\algname}{LMP}
\newcommand{\algnamepi}{LMP-$\pi$}
\newcommand{\algnamewam}{LMP-$\texttt{wam}$}
\newcommand{\algnametok}{LMP-\texttt{tok}}
\newcommand{\algnamefull}{Latent Memory Palace}
\newcommand{\std}[1]{\tiny{$\pm$ \makebox[10pt][r]{#1}}}
\newcommand{\bigstd}[1]{\tiny{$\pm$ \makebox[12pt][r]{#1}}}
\newcommand{\placeholder}{\makebox[14pt][c]{--}}

\newcommand{\cmt}[1]{{\footnotesize\textcolor{red}{#1}}}
\newcommand{\chuning}[1]{{\footnotesize\textcolor{blue}{Chuning: #1}}}
\newcommand{\abhishek}[1]{{\footnotesize\textcolor{cyan}{Abhishek: #1}}}
\newcommand{\paarth}[1]{{\footnotesize\textcolor{green}{Paarth: #1}}}
\newcommand{\jose}[1]{{\footnotesize\textcolor{purple}{Jose: #1}}}

\makeatletter
\newcommand{\onedot}{\ifx\@let@token.\else.\null\fi}
\makeatother
\def\eg{\emph{e.g}\onedot} \def\Eg{\emph{E.g}\onedot}
\def\ie{\emph{i.e}\onedot} \def\Ie{\emph{I.e}\onedot}
\def\cf{\emph{c.f}\onedot} \def\Cf{\emph{C.f}\onedot}
\def\etc{\emph{etc}\onedot} \def\vs{\emph{vs}\onedot}
\def\wrt{w.r.t\onedot} \def\dof{d.o.f\onedot}
\def\etal{\emph{et al}\onedot}
\makeatother

\newcommand{\eos}{\texttt{EOS}}
\newcommand{\kld}[2]{D_{\text{KL}}\bigl(#1 \parallel #2\bigr)}
\newcommand{\mi}[2]{I\bigl(#1;#2\bigr)}
\newcommand{\lstrict}{\prec}
\newcommand{\lestrict}{\preceq}

\newcommand{\prior}{$p_\theta(z|o)$}
\newcommand{\post}{$q_\theta(z|o, a)$}
\newcommand{\uncondprior}{$p(z)$}
\newcommand{\csobj}[2]{\mathrm{CS}\left(#1; #2\right)}

\newcommand{\ndim}{n_{\texttt{dim}}}
\newcommand{\nvocab}{n_{\texttt{vocab}}}

\newcommand{\sigmamax}{\sigma_{\text{max}}}
\newcommand{\sigmamin}{\sigma_{\text{min}}}

\maketitle

\begin{center}
    \vspace{-0.8cm}
    \url{https://weirdlabuw.github.io/lmp/}
\end{center}


\begin{abstract}
  Human decision-making is highly flexible -- some actions are taken immediately; others require longer deliberation. Language models have exhibited a similar capacity for adaptive ``reasoning.'' However, transferring this capability to continuous control policies has been challenging, as directly reasoning in language space may lack the granularity for spatial understanding and precise motions. In this work, we show that reasoning for control policies can emerge by organizing information in an autoregressive \emph{latent} space reminiscent of a memory palace, where retrieval is iterative and adaptive. Our method, \algnamefull{}~(\algname{}), formulates reasoning as \textit{variational inference with an autoregressive latent distribution}. We derive a latent-space reinforcement learning technique to tractably optimize its variational lower bound. The resulting policy, \algnamepi{}, achieves strong empirical performance in simulation and real-world domains while exhibiting interpretable, adaptive allocation of test-time compute. We further show that the same framework yields a variable-length action tokenizer, \algnametok{}, which significantly improves the performance of downstream autoregressive policies. Together, these results present a new perspective on latent reasoning for control through the lens of variational inference. 
\end{abstract}


\section{Introduction}

Human decision-making ranges from the reflexive (e.g. walking) to the deliberate (e.g. playing chess), with variability in the time and effort devoted to each decision. This process is both \emph{iterative} and \emph{adaptive}, proceeding through a chain of logical steps that scales with the complexity of the decision. Modern large language models (LLMs) exhibit a similar pattern: a broad class of ``reasoning'' LLMs achieves substantially improved performance by generating intermediate tokens before the final answer~\cite{wei2022cot, deepseekr1, openai2024openaio1card}.

We ask: can iterative, adaptive computation benefit sequential decision making problems such as robotics? For robotic policies, we posit that the benefits are twofold: efficient decision making under task variability, and improved generalization. However, directly transferring methods from language models is unlikely to suffice, as language tokens may not capture the nuances required for spatial understanding and precise motions~\cite{chen2024spatialvlmendowingvisionlanguagemodels}. We therefore propose that robotic policies reason iteratively in a \emph{latent space} learned end-to-end for action prediction. This preserves the benefits of iterative, adaptive computation while allowing the intermediate representations to capture control-relevant information at the appropriate granularity.

\begin{figure*}
    \centering
    \includegraphics[width=\linewidth]{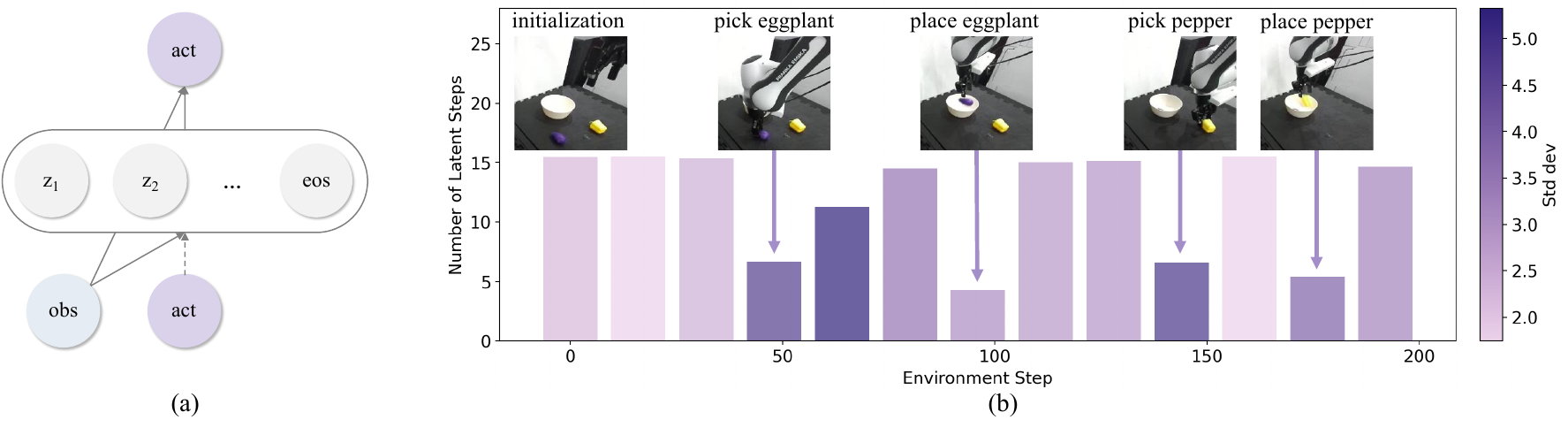}
    \caption{(a) \algnamefull{} (\algname{})  formulates iterative and adaptive reasoning as variational inference with a variable-length autoregressive latent distribution. (b) Applying \algname{} to decision-making results in control policies that adaptively allocate their test-time computation.}
    \label{fig:teaser}
    \vspace{-0.5cm}
\end{figure*}

To enable latent-space reasoning in robotic policies, we build on variational inference, a principled framework for learning latent-variable models by approximating posterior distributions over latent variables~\cite{kingma2013vae}. Applied to control, this perspective interprets an action not as a direct prediction from an observation, but as the outcome of an intermediate latent computation. However, standard variational inference typically uses fixed-dimensional latent variables, unable to adapt their computation to each input. To address this, we propose \algnamefull{} (\algname{}), a formulation of reasoning as \textit{variational inference with an autoregressive latent distribution}. As shown in Fig.~\ref{fig:teaser}(a), \algname{} encodes control-relevant information in an autoregressive latent space, where an \eos{} token allows information to be represented as variable-length paths. This structure resembles the memory palace, a mnemonic where information is associated with locations along a spatial route, and later recalled by mentally traversing that route one location at a time. In \algname{}, the autoregressive latent space is the palace, each latent token a location along a path, and generating the latent sequence amounts to traversing that path to retrieve the control-relevant information. Crucially, the \eos{} token lets the policy walk only as far as needed, giving rise to the iterative, adaptive retrieval that motivates our approach.

As in standard variational inference, we learn the latent tokens by maximizing a variational lower bound. This involves training a \emph{posterior} to encode each observation–action pair into a sequence of discrete latent tokens with two objectives: (i) reconstructing the action, and (ii) keeping the posterior close to a \emph{prior} conditioned on the observation alone. This prior, paired with the action decoder, precisely defines a policy \algnamepi{}. At test time, it generates latent ``reasoning'' tokens from the observation and decodes them into the action, adaptively allocating computation to each decision (Fig.~\ref{fig:teaser}(b)). To tractably optimize through the variable-length autoregressive sampling process, we convert the variational lower bound into a reinforcement learning (RL) objective over latent trajectories and apply RL techniques. We further show that by dropping the observation conditioning, the same framework instantiates a variable-length sequential action tokenizer, \algnametok{}, which compresses continuous actions into discrete latent tokens for training downstream autoregressive policies.

Experimentally, we show that \algnamepi{} yields control policies capable of iterative latent reasoning, with consistent performance gains across a range of robot learning tasks in both simulation and the real world. Our analyses indicate that (i) iterative computation outperforms non-iterative computation under the same variational framework, and (ii) parsimonious allocation of test-time compute, i.e. compression, plays an important role in the empirical performance. The action tokenizer, \algnametok{}, demonstrates substantial improvement over alternative tokenization schemes when evaluated with the same downstream autoregressive policy. Overall, these results establish autoregressive variational inference as a practical framework for iterative and adaptive computation in robot learning.
\section{Related Work}

\textbf{Reasoning for Foundation Models:} Reasoning has been shown to improve the output of foundation models via \emph{iterative inference} and \emph{adaptive test-time compute}. Chain-of-thought prompting improves accuracy in multi-step problems by generating intermediate steps~\citep{wei2022cot, yao2023treethoughtsdeliberateproblem}. RL-based reasoning models elicit and amortize iterative behaviors into inference~\citep{openai2024openaio1card, deepseekr1}. Test-time scaling laws show that allocating additional test-time compute can systematically improve model performance~\citep{snell2024scalingllmtesttimecompute, ma2025inferencetimescalingdiffusionmodels}. Latent reasoning shifts multi-step inference from explicit token-level traces to implicit representations, potentially increasing bandwidth and reducing latency~\citep{zhu2025surveylatentreasoning}. COCONUT progressively replaces explicit reasoning tokens with continuous embeddings via a training curriculum~\citep{coconut}. CODI replaces this curriculum with one-step distillation~\citep{shen-etal-2025-codi}. Extensions to multimodal settings similarly interleave latent visual representations with language tokens for visual reasoning~\citep{mirage}.
Most closely related is recent work on variational latent reasoning, which formulates reasoning as posterior inference over latent variables~\citep{zhou2025variationalreasoninglanguagemodels, Wang2026ReGuLaR, lin2025ravrreferenceanswerguidedvariationalreasoning}. Our work motivates latent reasoning through a different lens: robotic policies lack a natural reasoning medium. Therefore, reasoning in latent space provides a principled framework to bring the benefits of iterative and adaptive computation to robotics.

\textbf{Behavior Cloning via Generative Control Policies:} Behavior cloning (BC) is a standard approach for learning robot policies from demonstrations~\citep{alvinn}, increasingly cast as \emph{generative modeling} over actions. IBC represents action distributions with energy-based models to handle discontinuous or set-valued mappings~\citep{ibc}. Diffusion Policy models actions using iterative denoising processes, yielding expressive multimodal behavior with stable training~\citep{diffusionpolicy}. BeT and VQ-BeT enable multiple behavioral modes within transformer policies via discrete action tokens~\citep{bet, vqbet}. Our method is a discrete latent-variable model
capable of expanding computation as action prediction demands. As our experiments show, this approach achieves strong empirical performance while retaining the multimodal expressivity in prior generative control policies.

\textbf{Reasoning for Robotics:} Recent work brings iterative reasoning and test-time compute to embodied decision making through several complementary directions. First, some methods utilize \emph{explicit} intermediate reasoning traces to structure decision making: ECoT generates grounded sub-steps (e.g., sub-tasks) before acting~\citep{ecot}, ECoT Lite refines this recipe with a latency-aware design~\citep{ecotlite}, and OneTwoVLA combines fast reactive control with deliberative ``System~1/2'' computation~\citep{onetwovla}. Second, some works introduce \emph{structured intermediate representations} to bridge perception and control, such as spatial abstractions or hierarchical guidance~\citep{lee2025molmoactactionreasoningmodels, li2025hamsterhierarchicalactionmodels}. Concurrently, RD-VLA~\citep{tur2026rdvla} instantiates latent reasoning using a recurrent-depth transformer action head, halting when changes in predicted actions fall below a threshold. In contrast, our method reasons in an autoregressive latent space trained directly for action prediction, avoiding manually designed reasoning interfaces or stopping criteria.
\section{\algnamefull{}: Reasoning as Autoregressive Variational Inference}

We consider the imitation learning setting, where the goal is to learn a control policy $\pi_\theta: \mathcal{O} \rightarrow \mathcal{A}$ from a dataset of expert demonstrations, $\mathcal{D} = \{(o_i, a_i)\}_{i=1}^{n}$, with observations $o_i$ and actions $a_i$. For the sake of practicality, $o_i$ may be a stacked history of observations with language instructions, and $a_i$ may be a chunk of actions. The behavior cloning objective maximizes the likelihood of expert actions given observations: $J(\theta) = \mathbb E_{(o_i, a_i) \sim \mathcal D}\left[\log p_\theta (a_i|o_i)\right]$. We introduce a new parameterization of $\pi_\theta$ using a variable-length latent autoregressive model.

\subsection{Reasoning as Variational Inference}
\label{sec:ar-factorization}
We formulate reasoning for decision-making using a probabilistic latent variable framework. A natural way to introduce reasoning capabilities is to have the policy first output a latent reasoning ``trace'' $z$ that captures internal deliberation conditioned on the current observation $o$. This internal deliberation can then be used to output actions $a$ conditioned on $(o, z)$. This process can be formalized as a latent variable model (Fig. \ref{fig:teaser}(a)), where an action $a$ is generated from observation $o$ and latent variable $z$. Policy learning then amounts to maximum likelihood estimation of $p(a \mid o)$. Since this involves an intractable marginalization over $z$, we approximate the true posterior $p(z \mid o, a)$ with a variational distribution $q_\theta(z \mid o, a)$ and maximize the evidence lower bound (ELBO):
\begin{equation}
\log p(a \mid o)\ge \mathbb{E}_{z \sim q_\theta(z \mid o, a)}
\left[\log p_\phi(a \mid o, z)\right] - D_{\mathrm{KL}}\!\left(q_\theta(z \mid o, a)\,\|\,p_\theta(z \mid o)\right),
\label{eq:elbo1}
\end{equation}
where $\theta$ parametrizes the learned posterior and prior, and $\phi$ parametrizes the action decoder. \textit{The learned prior  distribution $p_\theta(z \mid o)$ and the decoder $p_\phi(a \mid o, z)$ fully parametrize a policy, where actions are generated by first sampling $z \sim p_\theta(z \mid o)$ and then sampling $a \sim p_\phi(a \mid o, z)$. }

However, this still leaves open the important question of what form the latent variable $z$ and the corresponding generative models should take. The transition from standard variational inference~\cite{kingma2013vae, vqbet} to \emph{latent reasoning} happens when we parameterize the latent space by a \emph{variable-length autoregressive model}. In particular, we will represent $z$ not as a single variable, but as a sequence of latent variables $z=(z_1,z_2,\ldots,z_{T(z)})$ produced from a vocabulary $\mathcal{V}\cup\{\eos\}$ with termination time
$T(z) \coloneqq \min\{t\ge 1 : z_t=\eos\}$. We parameterize the prior and the posterior with a factorized autoregressive family, $p_\theta(z \mid o) = \prod_{t=1}^{T(z)} p_\theta(z_t \mid z_{<t}, o)$, $q_\theta(z \mid o, a) = \prod_{t=1}^{T(z)} q_\theta(z_t \mid z_{<t}, o, a)$, and the likelihood with a full-sequence decoder $p_\phi(a \mid o, z) = p_\phi(a \mid o, z_{1:T(z)})$. This enables the model to iteratively generate sequences of latent tokens, terminating when an \eos{} token is generated. Together, the autoregressive factorization and the \eos{} token allow for \textit{iterative} and \textit{adaptive} computation. Under this factorization, Eq. \eqref{eq:elbo1} becomes:
\begin{equation}
\log p(a \mid o)
\;\ge\;
\mathbb{E}_{z_{1:T(z)} \sim q_\theta(\cdot \mid o,a)}
\left[
\log p_\phi(a \mid o, z_{1:T(z)})\right]
- D_{\mathrm{KL}}\!\left(
q_\theta(z_{1:T(z)} \mid o,a)\,\big\|\,p_\theta(z_{1:T(z)} \mid o)
\right)
.
\label{eq:elbo2}
\end{equation}

\paragraph{Adaptive Reasoning via Compression}
\label{sec:compression}
While the autoregressive factorization in principle allows for adaptive computation, Eq.~\eqref{eq:elbo2} does not specify when the computation should stop. To elicit adaptive behavior, we introduce compression so that the model only uses more steps as needed. We parameterize the decoder with an isotropic Gaussian distribution whose variance decays with latent length $p_\phi(a \mid o, z) = \mathcal N\left(\mu_\phi(o, z), \sigma^2(T(z))I\right)$, $\sigma(T(z)) = \gamma^{T(z)}\sigma_0$, $\gamma \in (0, 1)$. This effectively imposes a length penalty: each additional latent step sharpens the Gaussian, so the model can only afford another step if the predicted actions are precise enough (formalized in Appendix~\ref{app:derivation-stopping-time}). As a result, the model spends more compute when actions can be precisely predicted from the observation, and less when the remaining uncertainty is irreducible. For example, teleoperators close the gripper at slightly different times across demonstrations, so additional computation cannot pin down the exact timing. Empirically, we find that this design not only leads to adaptive compute allocation but also contributes significantly to performance (Sec.~\ref{exp:analysis}).
\begin{figure}
    \centering
    \includegraphics[width=1.0\linewidth]{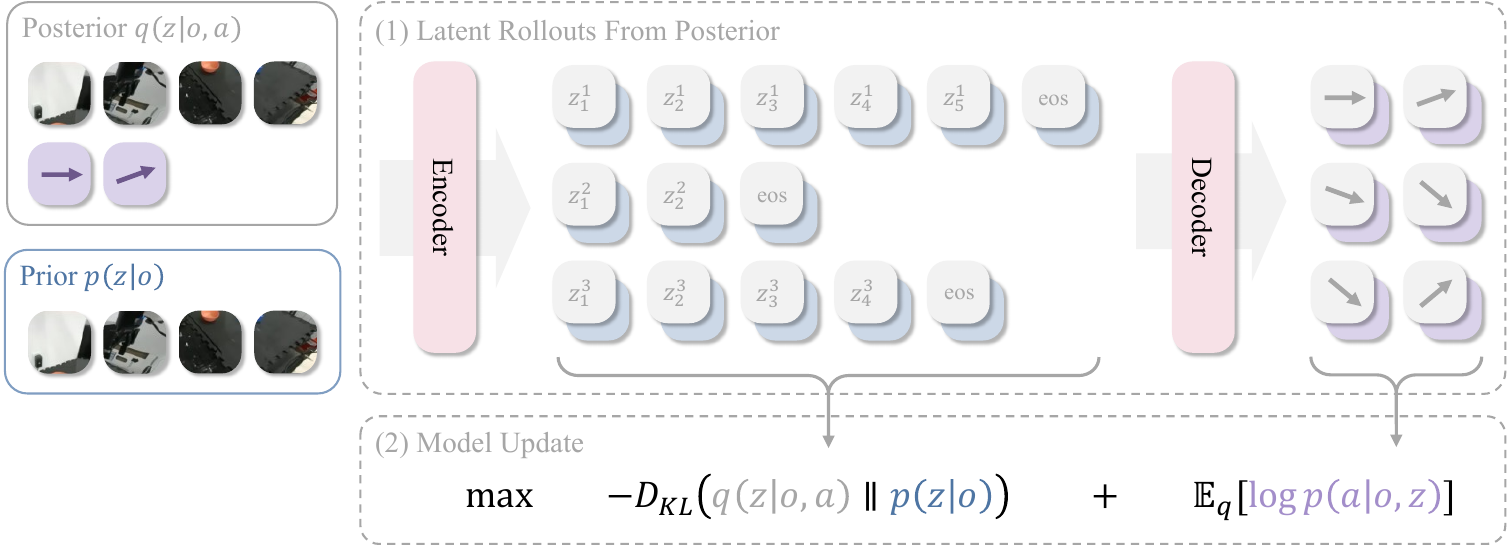}
    \caption{Training \algnamepi{} alternates between two stages. (1) \textbf{Latent rollout:} the posterior generates a buffer of latent trajectories. (2) \textbf{Model update:} the model optimizes a clipped-surrogate of the variational lower bound, computed from the posterior, the prior, and the decoder.}
    \label{fig:training}
    \vspace{-0.3cm}
\end{figure}

\subsection{Tractable Optimization via Latent-Space Reinforcement Learning}
Optimizing Eq.~\eqref{eq:elbo2} involves estimating the gradients through a non-differentiable autoregressive sampling process. To this end, we derive a tractable optimization objective using reinforcement learning. Note that  Eq.~\eqref{eq:elbo2} can be viewed as a reinforcement learning objective over latent-token trajectories.
Fixing $(o,a)$, define a latent ``episode'' that starts from the empty prefix and at each step samples a token $z_t$ from the \emph{variational posterior}
$q_\theta(z_t\mid z_{<t},o,a)$ until $\eos$.
This defines a variable-horizon rollout with ``policy'' $q_\theta(\cdot\mid o,a)$ and termination at $T(z)$.
A trajectory $z_{1:T(z)}$ is then scored by a return consisting of:
(i) a reconstruction reward $\log p_\phi(a\mid o,z_{1:T(z)})$, and
(ii) a penalty on the KL between the posterior $q_\theta(z_{1:T(z)} \mid o,a)$ and the prior $p_\theta(z_{1:T(z)} \mid o)$.
Note that the latent ``policy'' is different from the executable policy used to interact with the environment, which is sampled from
$z \sim p_\theta(\cdot\mid o), a \sim p_\phi(\cdot\mid o,z).$ \textit{RL is used only to optimize inference over latent ``reasoning'' chains, rather than interacting with the physical environment directly.}

\paragraph{Variance reduction via stepwise KL decomposition.} A naive policy-gradient estimator of Eq.~\eqref{eq:elbo2} has high variance because the return is only provided at the end of the rollout. We leverage the autoregressive structure of the KL term to decompose the episodic return into per-step rewards.
Let $\mathcal P_{\leq t} \coloneqq \{z_{\leq t}: z_i\neq \eos, \forall i<t\}$ denote the set of length-$t$ prefixes that have not terminated before step $t$. Let $q_\theta (z_{\leq t} \mid o, a) $ denote the prefix marginal mass over $z_{\leq t} \in \mathcal P_{\leq t}$, thus defining a subprobability measure whose total mass equals $\Pr_{q_\theta}[T(z) \geq t]$. Let $\tilde{\mathbb E}$ denote the expectation under this unnormalized measure.
As shown in Appendix \ref{app:derivation-per-step-kl}, the KL term admits the decomposition
\begin{equation}
\label{eq:step_kl}
D_{\mathrm{KL}}\!\left(q_\theta(z\mid o,a)\,\|\,p_\theta(z\mid o)\right)
=
\sum_{t=1}^{\infty}
\tilde{\mathbb{E}}_{z_{\leq t}\sim q_\theta(\cdot\mid o,a)}
\left[
\log \frac{q_\theta(z_t\mid z_{<t},o,a)}{p_\theta(z_t\mid z_{<t},o)}
\right].
\end{equation}

\paragraph{Optimization with a PPO-style clipped surrogate.}
\label{app:method-surrogate} 
To improve sample efficiency, we collect on-policy latent rollouts into a buffer and optimize the objective using a trust-region style surrogate~\cite{schulman2017ppo} as shown in Fig. \ref{fig:training}. Let $q_{\theta_{\mathrm{old}}}$ denote the stale rollout policy. Define likelihood ratio
$r(z) \coloneqq \frac{q_{\theta}(z\mid o,a)}{q_{\theta_{\mathrm{old}}}(z\mid o,a)}$, length-aware clipped likelihood ratio $\bar r(z) = \mathrm{clip}\!\left(r(z),\,(1-\epsilon)^{T(z)},\,(1+\epsilon)^{T(z)}\right)$, and clipped surrogate operator $\mathrm{CS}(r,\bar r; x) \;\coloneqq\; \min\!\left(r\,x,\; \bar r\,\mathrm{stopgrad}(x)\right)$
which updates $x$ only when $r$ remains inside the trust region.\footnote{Other equivalent PPO-style forms are possible; we found the above stable for variable-length rollouts.}
We maximize the surrogate objective
\begin{equation}
\label{eq:ppo_surrogate}
\begin{aligned}
J_{\mathrm{rec}}(\theta,\phi)
&=
\mathbb{E}_{z\sim q_{\theta_{\mathrm{old}}}(\cdot\mid o,a)}
\left[
\mathrm{CS}\left(r(z),\bar r(z);\; \log p_\phi(a\mid o,z)\right)
\right],\\
J_{\mathrm{kl}}^{t}(\theta)
&=\tilde{\mathbb{E}}_{z_{\leq t}\sim q_{\theta_{\mathrm{old}}}(\cdot\mid o,a)}
\left[\mathrm{CS}\left(
r(z_{\le t}),\bar r(z_{\le t});\;
\log \frac{p_\theta(z_t\mid z_{<t},o)}
{q_\theta(z_t\mid z_{<t},o,a)}\right)
\right],
\\
J(\theta,\phi)
&=
J_{\mathrm{rec}}(\theta,\phi)
+
\sum_{t=1}^{\infty} J_{\mathrm{kl}}^{t}(\theta).
\end{aligned}
\end{equation}
As in prior literature~\cite{higgins2017betavae}, we introduce additional stabilization techniques to prevent latent collapse and to balance the reconstruction and KL objectives. The details are presented in Appendix \ref{app:method-stabilization}.

\begin{figure}[t]
    \centering
    \includegraphics[width=1.0\linewidth]{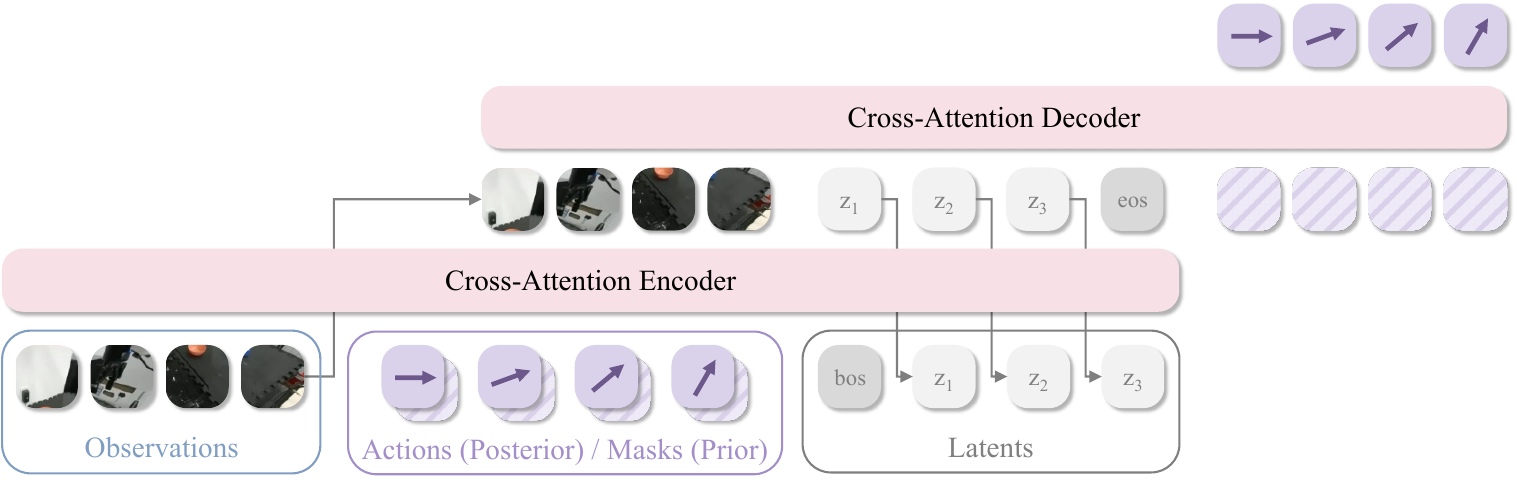}
    \caption{\algnamepi{} architecture.  The model consists of a latent encoder and an action decoder. The encoder is a causal transformer cross-attending to observations and actions. The decoder is a bidirectional transformer cross-attending to observations and latents.}
    \label{fig:model}
    \vspace{-0.3cm}
\end{figure}

\subsection{Connection to Action Tokenizer} In \algname{}, decisions are represented as variable-length paths through a discrete codebook. This representation naturally lends itself to the problem of action tokenization. In Appendix \ref{app:method-tokenizer}, we show that by dropping the observation conditioning from the variational inference framework, we can instantiate a variable-length sequential action tokenizer \algnametok{}. The tokenizer consists of an encoder $q_\theta(z_{1:T(z)} \mid a)$ which autoregressively generates tokens until \eos{}, and a decoder $p_\phi(a \mid z_{1:T(z)})$ which converts token sequences into continuous actions. \algnametok{} can be used as the output space of downstream autoregressive models for learning expressive policies. Note that with \algnametok{}, a latent sequence of arbitrary length can be decoded into valid actions. This stands in contrast with existing tokenizers like FAST~\cite{pertsch2025fast} which requires the output tokens to exactly reconstruct a discrete cosine transform (DCT) matrix, invalidating partial token sequences.

\subsection{Practical Instantiation}
We describe a practical instantiation of \algnamepi{} and \algnametok{}. As shown in Fig. \ref{fig:model}, we parameterize the prior \prior{} and the posterior \post{} using a shared causal transformer with cross-attention conditioning. The posterior \post{} is conditioned on image observation embeddings and action embeddings. The prior \prior{} is only conditioned on the observation embeddings, masking out actions with learned mask tokens. The action decoder is a bidirectional transformer conditioned on the observation embeddings and the sampled latent tokens, outputting action chunks from learned embeddings. During training, we truncate the latent sequences at a fixed length $H$ with shorter sequences padded to the same length. For \algnamepi{}, we define the decoder variance schedule (Sec.~\ref{sec:compression}) by specifying $(\sigmamax, \sigmamin)$ with an exponential decay. We additionally let the decoder predict a per-dimension scale $s \in [0, 1]$ applied on top of the variance schedule, so that the model can commit to small variance even when terminating early. For \algnametok{}, we use a scalar learned variance applied to all timesteps, similar to $\sigma$-VAE~\cite{sigmavae}.

\section{Experiments}
\label{sec:experiments}
\begin{figure}[t]
    \centering
    \includegraphics[width=1.0\linewidth]{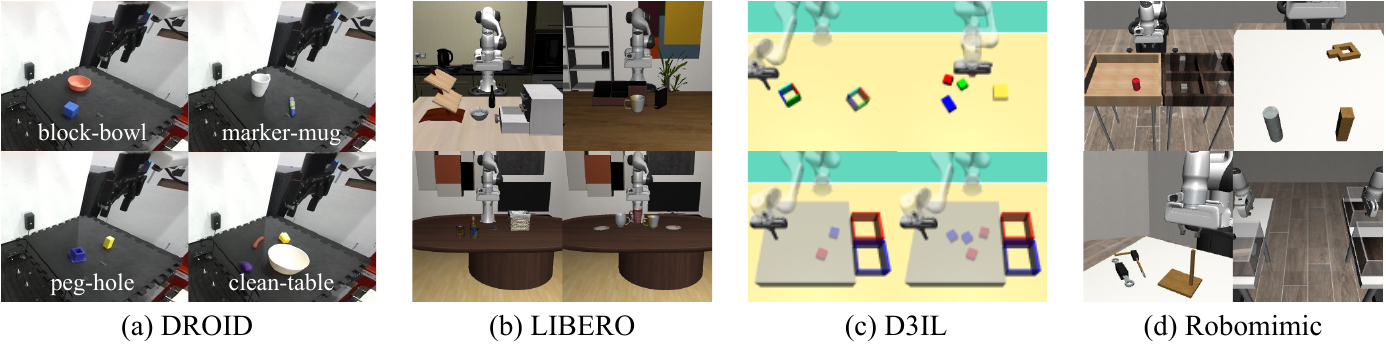}
    \vspace{-0.3cm}
    \caption{\footnotesize{Visualization of evaluation domains.}}
    \label{fig:environments}
    \vspace{-0.3cm}
\end{figure}
We conduct experiments to answer the following research questions. (1) Is \algnamepi{} a performant generative control policy? (2) Is \algnametok{} an effective action tokenizer? (3) Does \algname{} demonstrate adaptive allocation of test-time compute? (4) Which factors contribute to \algname{}'s performance?

\paragraph{Environments}
We evaluate our method on real-world and simulated visuomotor control benchmarks as shown in Fig. \ref{fig:environments}. (1) DROID~\cite{droid} is a real-world manipulation platform with multitask training data. We evaluate language-conditioned multitask policies in zero-shot and finetuned settings. For each task, we evaluate on 20 randomly sampled initial conditions. (2) LIBERO~\cite{liu2023libero} is a simulated manipulation benchmark featuring diverse task scenarios.  We evaluate language-conditioned multitask policies in-distribution. (3) D3IL~\cite{jia2024d3il} is a simulated benchmark emphasizing multimodality. (4) RoboMimic~\cite{mandlekar2021robomimic} is a simulated benchmark featuring high-precision tasks. Additional environment details are provided in Appendix \ref{app:envs}. All simulation results are averaged over three seeds.

\begin{table}[t]
\centering
\small
\caption{Evaluation Results Across DROID Tasks.}
\begin{tabular}{lcccccc}
\toprule
 & \multicolumn{2}{c}{zero-shot} & \multicolumn{4}{c}{finetuned} \\
\cmidrule(lr){2-3} \cmidrule(lr){4-7}
& block-bowl & marker-mug & peg-hole & clean-table (1/3) & clean-table (2/3) & clean-table (3/3) \\
\midrule
DP & 0.40 & 0.25 & 0.30 & 0.80 & 0.55 & 0.25 \\
\algnamepi{} & \textbf{0.65} & \textbf{0.55} & \textbf{0.70} & \textbf{0.95} & \textbf{0.95} & \textbf{0.55}\\
\bottomrule
\end{tabular}
\label{tab:exp-policy-real}
\vspace{-0.3cm}
\end{table}

\subsection{Is \algnamepi{} a performant generative control policy?} 
\label{exp:policy}
We compare \algnamepi{} against Diffusion Policy (DP)~\cite{diffusionpolicy}, a state-of-the-art generative control policy parameterized by a diffusion transformer similar to Large Behavior Models~\cite{lbmtri2025}. We instantiate both methods as 1B-parameter, language-conditioned policies. Table \ref{tab:exp-policy-real} shows the results on the real-world DROID platform. In the zero-shot setting, \algnamepi{} outperforms DP on block-bowl and marker-mug, indicating stronger out-of-the-box generalization. We then finetune the policies on two challenging tasks, peg-hole and clean-table, designed to measure precision and compositional generalization respectively. We find \algnamepi{} to outperform DP on both tasks, notably picking up at least 2 objects with 95\% success rate in the clean-table task. These results suggest that \algnamepi{} is capable of precise control and compositional generalization. Table \ref{tab:exp-policy-sim} shows the language-conditioned multitask results on LIBERO-90. \begin{wraptable}{r}{0.4\linewidth}
\centering
\small
\caption{Evaluation Results Across LIBERO-90 Tasks.}
\begin{tabular}{lcc}
\toprule
& all tasks & bottom-10 tasks \\
\midrule
DP & 0.909 \bigstd{0.004} & 0.463 \bigstd{0.011} \\
\algnamepi{} & \textbf{0.933 \bigstd{0.003}} & \textbf{0.645 \bigstd{0.023}} \\
\bottomrule
\end{tabular}
\label{tab:exp-policy-sim}
\end{wraptable}While \algnamepi{} performs marginally better in terms of overall success rate, a closer examination of the bottom 10 tasks of each method reveals that \algnamepi{} has a significantly higher floor than DP. This pattern indicates reduced cross-task interference and better retrieval. Fig.~\ref{fig:per-task-success} provides a detailed comparison of per-task success rates, where we see \algnamepi{} tapering off more slowly than DP.

\subsection{Is \algnametok{} an effective action tokenizer?}
\label{exp:tokenizer}
\begin{wrapfigure}{r}{0.4\linewidth}
    \vspace{-0.5cm}
    \centering
    \includegraphics[width=\linewidth]{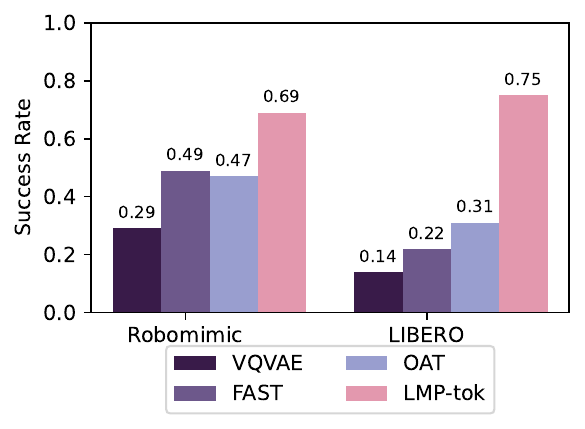}
    \caption{Average success rates of autoregressive policies fitted to tokenizers on Robomimic and LIBERO tasks.}
    \vspace{-0.3cm}
    \label{fig:results-tokenizer}
\end{wrapfigure}
We compare \algnametok{} with three representative action tokenizer baselines: (1) VQ-VAE~\cite{lee2024vqbet}, a 2-layer residual VQ tokenizer, (2) FAST~\cite{pertsch2025fast}, a variable-length action tokenizer based on DCT transform and byte-pair encoding, and (3) OAT~\cite{liu2026orderedactiontokenization}, a variable-length latent action tokenizer with a coarse-to-fine inductive bias. We evaluate the methods on RoboMimic and LIBERO, each featuring a large pretraining action dataset and multiple downstream tasks. We apply the tokenizers to the same downstream autoregressive transformer policy (with an added \eos{} token for variable-length methods). As shown in Fig. \ref{fig:results-tokenizer}, \algnametok{} significantly outperforms baselines in the downstream tasks. Moreover, individual task results in Table \ref{tab:exp-tokenizer} show that \algnametok{} is the only method achieving nontrivial performance on the high-precision RoboMimic tool-hang task. VQ-VAE lacks sufficient precision for fine-grained manipulation. FAST requires exact reconstruction of DCT matrices, placing a heavy burden on the downstream policy. While OAT belongs to the same class of latent autoregressive action tokenizers, we find it less effective than \algnametok{}, likely because it is trained with only reconstruction and not a variational objective to maintain latent diversity. 

\subsection{Analysis}
\label{exp:analysis}
We conduct analytical experiments to ablate the factors that contribute to \algname{}'s performance, and to understand the mechanism behind iterative and adaptive reasoning. We provide additional ablation results in Appendix \ref{app:exps}.

\begin{wrapfigure}{r}{0.4\textwidth}
    \vspace{-0.5cm}
    \centering
    \includegraphics[width=\linewidth]{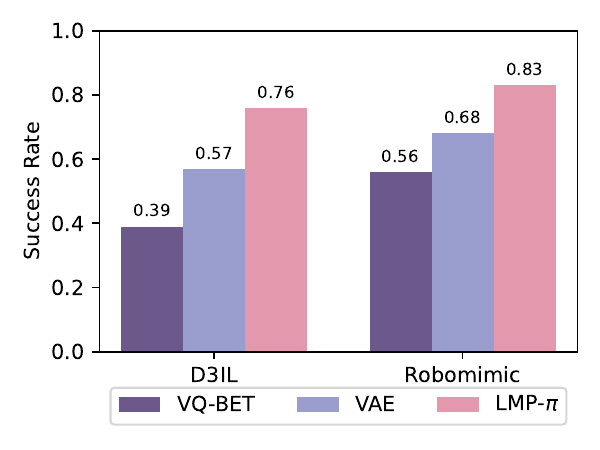}
    \caption{Average success rates of single-task policies on D3IL and RoboMimic.}
    \label{fig:results-singletask}
    \vspace{-0.3cm}
\end{wrapfigure}
\textbf{Does iterative computation benefit over non-iterative computation?}
\label{exp:non-iterative}
We show the benefit of iterative computation by comparing \algnamepi{} to two non-iterative latent-variable baselines: (1) VAE Policy~\cite{kingma2013vae}, which fits a one-step Gaussian VAE to expert actions conditioned on observations, and (2) VQ-BeT~\cite{lee2024vqbet}, which combines a behavior transformer with a residual VQ-VAE tokenizer. Although VQ-BeT uses an autoregressive transformer, its two-level hierarchy makes it effectively non-iterative. As shown in Fig. \ref{fig:results-singletask}, \algname{} consistently outperforms non-iterative latent-variable models on both D3IL and RoboMimic tasks, indicating that fixed-length latent representations lack the granularity needed for complex manipulation tasks.

\begin{figure}
    \centering
    \includegraphics[width=1.0\linewidth]{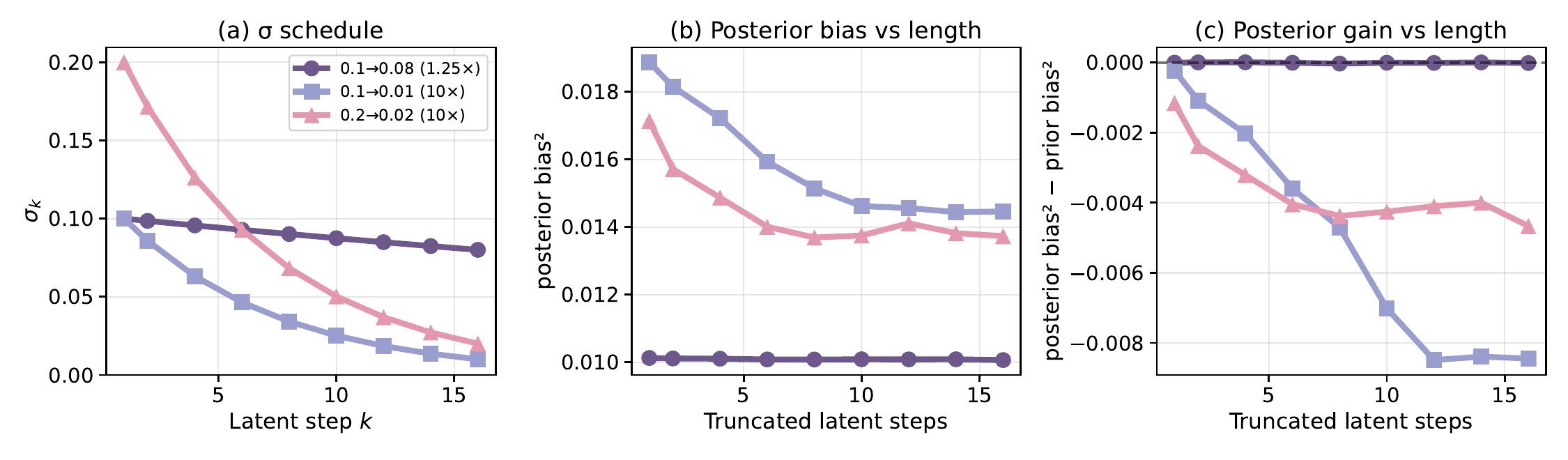}
    \caption{Effect of compression on latent utilization. (a) $\sigma$ schedules defined by $\sigmamax$-$\sigmamin$. (b)-(c) reconstruction error vs number of truncated latent steps. With lower compression (0.1-0.08 vs 0.1-0.01), the model ignores the latents, as indicated by the flat posterior error and posterior gain. Higher compression (0.1-0.01 vs 0.2-0.02) leads to increased error. We find a balanced configuration (0.2-0.02) to work well empirically.}
    \label{fig:analysis-compression}
\end{figure}

\begin{wraptable}{r}{0.4\linewidth}
\centering
\small
\caption{Ablation of Variance Schedule. Lower $\sigmamin$ means higher compression. $\sigmamax$ is fixed at 0.1.}
\begin{tabular}{lcc}
\toprule
$\sigmamin$ & all tasks & bottom-10 tasks \\
\midrule
0.01 & \textbf{0.933 \bigstd{0.003}} & \textbf{0.645 \bigstd{0.023}} \\
0.03 & 0.913 \bigstd{0.001} & 0.567 \bigstd{0.006} \\
0.05 & 0.870 \bigstd{0.012} & 0.450 \bigstd{0.045} \\
\bottomrule
\end{tabular}
\label{tab:exp-ablate-variance}
\end{wraptable}
\textbf{Does compression contribute to performance?} 
To analyze the effect of compression on policy performance, we ablate the compression strength by controlling the variance schedule. Recall from Sec. \ref{sec:compression} that the lower the $\sigmamin$, the faster the variance decays, and the stronger the compression. Table \ref{tab:exp-ablate-variance} shows LIBERO-90 results with different levels of compression. We find that stronger compression (lower $\sigmamin$) leads to improved policy performance. This suggests that parsimonious allocation of latent compute is important for generalization. Fig. \ref{fig:analysis-compression} analyzes the effect of compression on reconstruction error and latent utilization using DROID checkpoints. We find that (1) insufficient compression leads to latents being ignored by the decoder (0.1-0.08 vs. 0.1-0.01), and (2) excessive compression forces the model to use shorter traces and thus raises the overall reconstruction error (0.2-0.02 vs. 0.1-0.01). Empirically, we find a configuration that balances reconstruction and latent utilization (0.2-0.02) to work well. 

\begin{figure*}[t]
    \centering
    \includegraphics[width=1.0\linewidth]{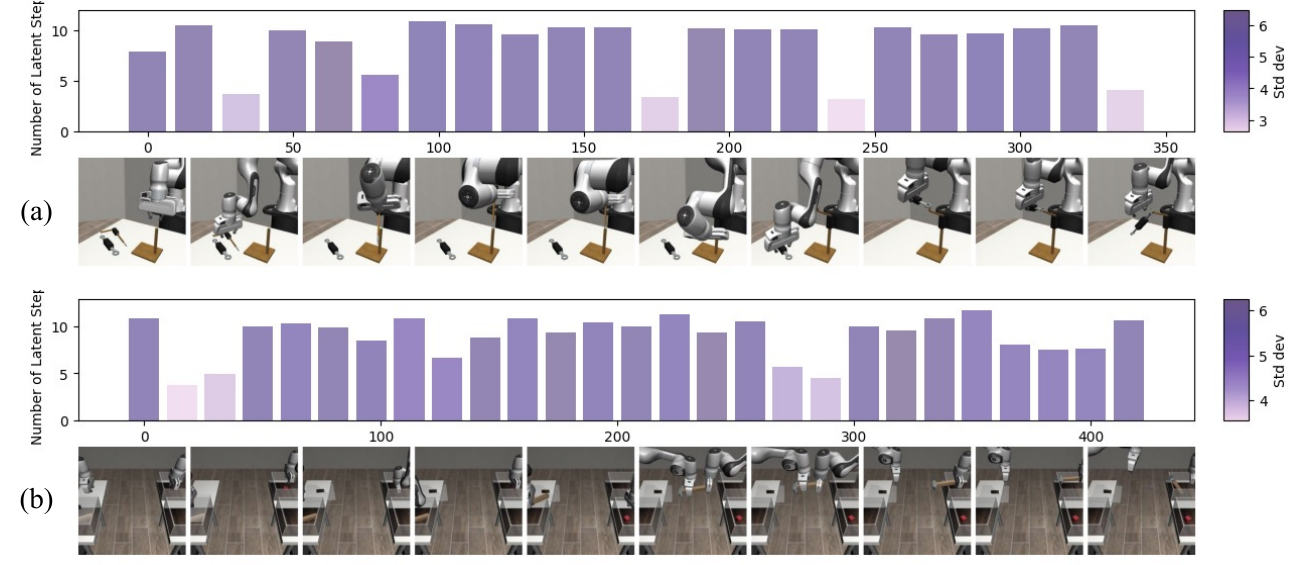}
    \caption{\algnamepi{} exhibits adaptive and interpretable allocation of latent steps. The length of latent traces is lower during gripper movements, e.g. (a) grasping and releasing, and (b) hand-over. }
    \label{fig:vis-reasoning-steps}
    \vspace{-0.3cm}
\end{figure*}

\paragraph{Qualitative analysis of test-time compute} 
\begin{wrapfigure}{r}{0.4\linewidth}
    \centering
    \includegraphics[width=\linewidth]{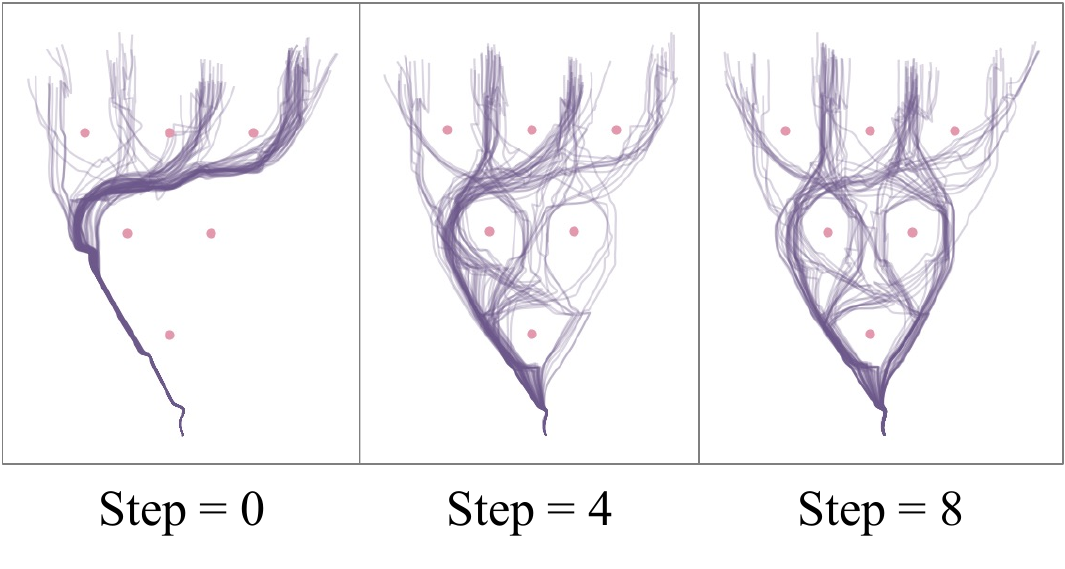}
    \caption{As the number of latent steps is truncated, the policy exhibits mode-seeking behavior.}
    \label{fig:ablation-multimodality}
\end{wrapfigure}
In Fig. \ref{fig:vis-reasoning-steps}, we visualize the number of latent steps generated by \algnamepi{} during rollouts. To reduce variance, we generate 64 latent traces for each observation and plot the average number of reasoning steps. We find that the model uses fewer reasoning steps when the robot must utilize the gripper. For example, in the DROID clean-table task (Fig. \ref{fig:teaser}(b)), fewer latent steps are used during grasping and releasing. Similar patterns appear in RoboMimic tool-hang and transport (Fig. \ref{fig:vis-reasoning-steps}), as well as other simulated and real-world environments (Appendix \ref{app:exps}). As discussed in Sec. \ref{sec:compression}, this pattern reflects the irreducible timing variability of gripper commands. 
Fig. \ref{fig:ablation-multimodality} shows a qualitative analysis of reasoning steps and multimodality. We train \algnamepi{} on the d3il-avoiding task and plot its rollouts as we manually truncate the maximum number of latent steps at test time. We find that the model effectively becomes deterministic as the number of latent steps decreases to zero (barring environment stochasticity). Notably, despite the model being deterministic at low latent steps, it exhibits mode-seeking behavior and does not suffer from the typical mode-averaging issue of deterministic policies. We hypothesize that this behavior is a result of training the model with RL rather than supervised learning. 

\paragraph{Quantitative analysis of test-time compute}
\begin{wrapfigure}{r}{0.4\linewidth}
    \vspace{-0.2cm}
    \centering
    \includegraphics[width=\linewidth]{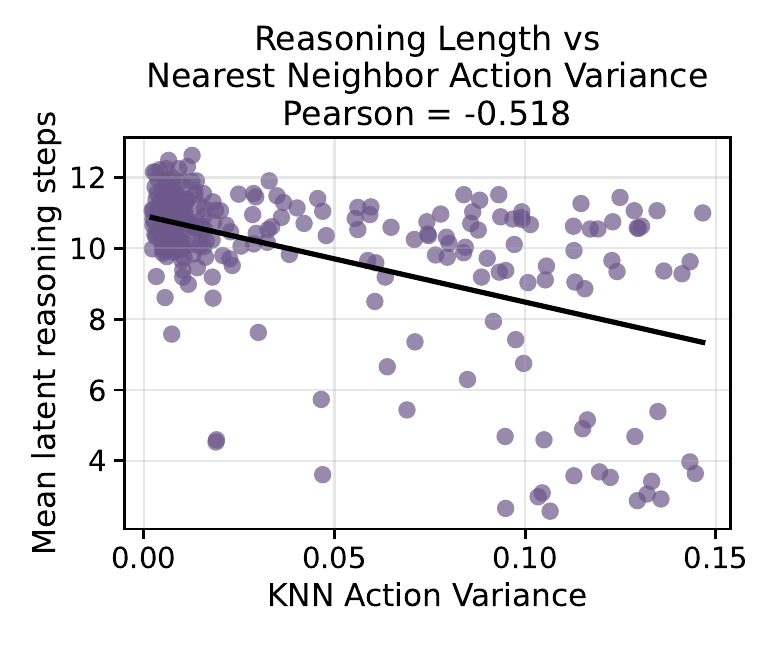}
    \caption{Average latent step vs variance of KNN actions. The length of reasoning steps is negatively correlated with variance of the actions at the $k=32$ nearest states in the dataset. }
    \label{fig:analysis-knn}
    \vspace{-0.2cm}
\end{wrapfigure}
In Fig. \ref{fig:analysis-knn}, we establish a correlation between the number of reasoning steps and action uncertainty. We roll out a LIBERO multitask \algnamepi{} on the ``stack right bowl on left bowl and put them in drawer'' task for 20 episodes. At each state, we query the task-specific dataset for the $k=32$ nearest-neighbor states and compute the variance of the actions taken at these nearest neighbors. This provides a nonparametric estimate of action uncertainty. Results in Fig. \ref{fig:analysis-knn} show a negative correlation between the number of reasoning steps and the KNN action variance ($r=-0.518$). This means the model uses fewer steps when the expert actions near the current state have high variance, i.e. when there is more irreducible uncertainty. We find most of the variance to be explained by binarized gripper actions whose timing varies across demonstrations. This results in the model spending fewer steps during gripper movements. 
\section{Discussion}
We introduce \algnamefull{}, a formulation of iterative and adaptive reasoning as variational inference with an autoregressive latent distribution. Applied to sequential decision making, this formulation yields a policy class, \algnamepi{}, which exhibits strong empirical performance and adaptive allocation of test-time compute. Our analyses show that both iterative computation and adaptive compute allocation are essential to performance. We also introduce a variable-length action tokenizer, \algnametok{}, which significantly outperforms existing methods when controlling for the downstream policy. These results establish autoregressive variational inference as a principled and practical route to iterative, adaptive computation for robotic control. \algname{} opens up several future directions. One is to replace the discrete autoregressive distribution with a continuous chain of Gaussians resembling a diffusion process. Another is to make latents persistent across an episode, rather than regenerating them from scratch at each step.

\paragraph{Limitations}
Since \algname{} is trained with sampling-based RL techniques, it is sensitive to hyperparameters. For example, the latent distribution is prone to collapsing without a sufficiently large rollout buffer or regularization. We leave the exploration of stable and scalable optimization of the proposed framework to future work.


\clearpage
\acknowledgments{This work is supported by TRI under the U3.0 program. Chuning Zhu is supported by the Amazon Core AI Fellowship. We thank Kushal Aurora, Yiqing Xu, Sriyash Poddar, Marius Memmel, Jesse Zhang, Patrick Yin, Arhan Jain, Jacob Berg, Pranav Teegavarapu and members of the WEIRD lab for helpful discussions.}


\bibliography{references}  

\clearpage
\appendix
\section{Extended Method Details}
\label{app:method}

\subsection{Posterior stabilization}
\label{app:method-stabilization}
Optimizing Eq. \eqref{eq:ppo_surrogate} involves jointly training the prior $p_\theta(z\mid o)$ and the posterior $q_\theta(z\mid o, a)$, which is prone to collapse. We mitigate this with two  stabilizers.

\emph{(i) Uniform-interpolated regularization.}
We regularize the posterior toward a length-aware uniform prior $u(z_t)$ over $\mathcal{V}\cup\{\eos\}$. We define a geometrically interpolated prior at each step:
$$
p_{\mathrm{mix}}^{\alpha}(z_t\mid z_{<t},o)
\;\propto\;
p_\theta(z_t\mid z_{<t},o)^{\alpha}\, u(z_t)^{1-\alpha},
\hspace{0.1cm} \alpha\in[0,1],
$$
and replace $p_\theta$ with $p_{\mathrm{mix}}^\alpha$ in the KL term:
\begin{equation}
\label{eq:kl_mix}
\begin{aligned}
J_{\mathrm{kl}}^{t}(\theta,\alpha)
=
\tilde{\mathbb{E}}_{z_{\leq t}\sim q_{\theta_{\mathrm{old}}}(\cdot\mid o,a)}
\left[
\mathrm{CS}\left(r(z_{\leq t}),\bar r(z_{\leq t});\;
\log \tfrac{p_{\mathrm{mix}}^\alpha(z_t\mid z_{<t},o)}{q_\theta(z_t\mid z_{<t},o,a)}
\right)
\right].
\end{aligned}
\end{equation}

\emph{(ii) Per-step free-nats clipping.}
We cap the strength of the KL ``pull'' toward the prior at each step using a free-nats threshold, preventing overly aggressive distillation early in training.
Let $\rho_t \coloneqq \Pr_{q_\theta}[T(z)\geq t]$ denote the survival probability and $D_{\mathrm{KL}}^t \coloneqq \kld{q_\theta(z_t\mid z_{<t},o,a)}{p_\theta(z_t\mid z_{<t},o)}$ denote the one-step KL. The per-step KL term in Eq. \ref{eq:step_kl} can be rewritten as survival-weighted expected one-step KL:
\begin{equation}
\tilde{\mathbb{E}}_{z_{\leq t}\sim q_\theta(\cdot\mid o,a)}\left[
\log \frac{q_\theta(z_t\mid z_{<t},o,a)}{p_\theta(z_t\mid z_{<t},o)}
\right] = \rho_t\,\mathbb E_{z_{<t}\sim q_\theta(\cdot\mid o,a,\,T(z)\geq t)}\left[D_{\mathrm{KL}}^t \right].
\end{equation}
We clip the expected one-step KL $\mathbb E_{z_{<t}\sim q_\theta(\cdot\mid o,a,\,T(z)\geq t)}\left[D_{\mathrm{KL}}^t \right]$ by a free nat budget $\tau \geq 0$. Applying this to the clipped surrogate objective gives:
\begin{equation}
\label{eq:free_nats}
\begin{aligned}
J_{\mathrm{kl}}^{t}(\theta,\alpha, \tau)
=
\rho_t \min\left(-\tau, \mathbb E_{z_{\leq t} \sim q_{\theta_\mathrm{old}} (\cdot \mid o, a, T(z) \geq t)}
\left[
\mathrm{CS}\left(r(z_{\leq t}),\bar r(z_{\leq t});\;
\log \tfrac{p_{\mathrm{mix}}^\alpha(z_t\mid z_{<t},o)}{q_\theta(z_t\mid z_{<t},o,a)}
\right)
\right]\right).
\end{aligned}
\end{equation}
This effectively applies a soft constraint to the KL contribution at each timestep. In practice, we set $\tau=\beta\log| \mathcal{V} \cup \{\eos\}|, \beta \in [0, 1]$, to scale free nats proportionally to the maximum entropy of a categorical variable.

\subsection{Tokenizer objective}
\label{app:method-tokenizer}
The tokenizer objective is similar to the policy (Eq. \ref{eq:elbo2}) but drops the observation conditioning. Let $q_\theta(z_{1:T(z)} \mid a)$ denote the learned posterior and $p(z_{1:T(z)})$ denote a prior where each $p(z_t \mid z_{<t})$ is uniform over $\mathcal V \cup \{\eos\}$. \algnametok{} maximizes the ELBO:
\begin{equation}
\log p(a)
\;\ge\;
\mathbb{E}_{z_{1:T(z)} \sim q_\theta(\cdot \mid a)}
\left[
\log p_\phi(a \mid z_{1:T(z)})\right]
- D_{\mathrm{KL}}\!\left(
q_\theta(z_{1:T(z)} \mid a)\,\big\|\,p(z_{1:T(z)})
\right)
.
\label{eq:elbo_tok}
\end{equation}

The clipped surrogate objective is:
\begin{equation}
\begin{aligned}
J_{\text{rec}}(\theta, \phi) &= 
\mathbb E_{q^{\text{old}}_\theta(z \mid a)} \left[\mathrm{CS}\left(r(z), \bar r(z); \log p_\phi(a\mid z) \right)\right], \\
J_{\text{kl}}^t(\theta) &= 
\tilde{\mathbb E}_{q^{\text{old}}_\theta(z_{\leq t}\mid a)} \left[ \mathrm{CS}\left(r(z_{\leq t}), \bar r(z_{\leq t}); \log \frac{p(z_t)}{q_\theta(z_t\mid z_{< t}, a)}\right)\right], \\
J(\theta, \phi) &= J_{\text{rec}}(\theta, \phi) + \sum_{t=1}^{\infty}J^t_{\text{kl}}(\theta).
\label{eq:ppo_surrogate_tokenizer}
\end{aligned}
\end{equation}

\subsection{Instantiation as World Action Model}
We can extend \algnamepi{} to a world action model~\cite{zhu2025uwm, ye2026worldactionmodelszeroshot} by noticing that under deterministic and invertible dynamics, actions and future observations are equivalent given current observations: $(o_{t}, a_{t}) \sim (o_{t}, o_{t+1})$. This allows us to instantiate a world action model, \algnamewam{}, by replacing the encoder's input actions $a_t$ with future observations $o_{t+1}$. The model jointly predicts future observations in the latent space (via the KL objective) and actions explicitly (via the reconstruction objective). In Table \ref{tab:exp-policy-sim-full}, we show that this formulation achieves comparable performance to \algnamepi{} on the LIBERO-90 multitask benchmark. We leave exploration of the world action model formulation to future work.
\begin{table}[t]
\centering
\small
\caption{Full Evaluation Results Across LIBERO-90 Tasks.}
\begin{tabular}{lccc}
\toprule
 & DP & \algnamepi{} & \algnamewam{} \\
\midrule
all tasks           & 0.909 \bigstd{0.004}  & \textbf{0.933 \bigstd{0.003}}  & \textbf{0.928 \bigstd{0.002}} \\
bottom-10 tasks     & 0.463 \bigstd{0.011}  & \textbf{0.645 \bigstd{0.023}} & \textbf{0.635 \bigstd{0.005}} \\
\bottomrule
\end{tabular}
\label{tab:exp-policy-sim-full}
\end{table}
\section{Derivations}
\subsection{Derivation of Per-Step KL Objective}
\label{app:derivation-per-step-kl}
Optimizing $\kld{q(z \mid o, a)}{p(z \mid o)}$ in Eq.~\ref{eq:elbo2} is subject to high variance. To stabilize training, we decompose the sequence-level KL into per-step contributions, effectively breaking an episodic return into per-step rewards. For clarity, we omit the conditioning on $(o, a)$. 

Assume (i) $q(z)$ and $p(z)$ are defined over finite \eos-terminated sequences, i.e. $\Pr_q[T(z) < \infty] = 1$ and $\Pr_p[T(z) < \infty] = 1$, (ii) $q(z)$ and $p(z)$ have compatible support, i.e. $q(z)>0 \Rightarrow p(z)>0$, and (iii) the sequence-level KL is finite, i.e. $\kld{q(z)}{p(z)} < \infty$. We first carry out the derivation on a finite horizon $H$ and then pass to the limit. To compare sequences of different lengths, we extend each sequence that has terminated by step $H$ with deterministic null tokens after \eos, where the padded positions satisfy $q(z_t\mid z_{<t}) = p(z_t\mid z_{<t}) = 1$ and contribute zero log-ratio. Writing $z_{\leq H}$ for the padded length-$H$ prefix and applying the autoregressive factorization, we have:
\begin{align*}
    \kld{q(z_{\leq H})}{p(z_{\leq H})} &= \sum_{z_{\leq H}} q(z_{\leq H}) \log \frac{q(z_{\leq H})}{p(z_{\leq H})} \\
    &= \sum_{z_{\leq H}} q(z_{\leq H}) \sum_{t=1}^H \log \frac{q(z_t|z_{<t})}{p(z_t|z_{<t})} \\
    &= \sum_{t=1}^H \sum_{z_{\leq H}} q(z_{\leq H}) \log \frac{q(z_t|z_{<t})}{p(z_t|z_{<t})} \quad\text{(exchanging finite sums).}
\end{align*}
Since the padded positions after \eos{} contribute zero log-ratio, only prefixes that have not yet terminated remain. Let $\mathcal P_{<t} \coloneqq \{z_{<t}: z_i\neq \eos, \forall i<t\}$ denote the set of length-$(t-1)$ prefixes that have not terminated before step $t$. We have:
\begin{align*}
    \kld{q(z_{\leq H})}{p(z_{\leq H})}
    &= \sum_{t=1}^H \sum_{z_{<t} \in \mathcal P_{<t}} \sum_{z_t} q(z_{<t})\, q(z_t\mid z_{<t}) \log \frac{q(z_t\mid z_{<t})}{p(z_t\mid z_{<t})} \\
    &= \sum_{t=1}^H \sum_{z_{<t} \in \mathcal P_{<t}} q(z_{<t}) \kld{q(z_t\mid z_{<t})}{p(z_t\mid z_{<t})}.
\end{align*}
This expresses the finite-horizon KL as a sum of nonnegative per-step terms. By assumption (i), the length-$H$ prefix determines the full sequence in the limit, so the left hand side converges to $\kld{q(z)}{p(z)}$. The right hand side is a series of nonnegative per-step partial sums, whose limit exists by monotone convergence. Taking the limit of both sides gives:
\begin{align*}
    \kld{q(z)}{p(z)}
    = \sum_{t=1}^\infty \sum_{z_{<t} \in \mathcal P_{<t}} q(z_{<t}) \kld{q(z_t\mid z_{<t})}{p(z_t\mid z_{<t})}.
\end{align*}
Let $\mathcal P_{\leq t} \coloneqq \{z_{\leq t}: z_i\neq \eos, \forall i<t\}$ denote the set of length-$t$ prefixes that have not terminated before step $t$. Let $q(z_{\leq t})$ denote the marginalized prefix mass over $z_{\leq t} \in \mathcal P_{\leq t}$, thus defining a subprobability measure whose total mass equals $\Pr_q[T(z) \geq t]$. Let $\tilde{\mathbb E}$ denote the expectation under this unnormalized measure. The expression above can be equivalently written as:
\begin{align*}
    \kld{q(z)}{p(z)} &= \sum_{t=1}^{\infty} \tilde{\mathbb E}_{z_{\leq t}\sim q(\cdot)} \left[\log \frac{q(z_t\mid z_{<t})} {p(z_t\mid z_{<t})}\right].
\end{align*}

\subsection{Derivation of Stopping Criterion}
\label{app:derivation-stopping-time}
We analyze the stopping criterion for the autoregressive process under the decoder variance schedule from Sec. \ref{sec:compression}. We isolate the reconstruction objective and omit the KL term to provide a heuristic understanding of the latent dynamics. Let $D$ denote the action dimension, $\mu_L:=\mu_\phi(o, z_{1:L})$ the decoder mean at length $L$, $\varepsilon_L := ||a - \mu_L||^2 / D$ the per-dimension decoder mean squared error at length $L$, and $\sigma(L)=\gamma^L\sigma_0$ the decoder standard deviation with discount $\gamma \in (0, 1)$.  The action log-likelihoods at length $L$ and $L+1$ are:
\begin{align*}
    \ell(L) &= -\frac{D}{2}\log(2\pi) - D\log\sigma(L) - \frac{D\,\varepsilon_L}{2\,\sigma^2(L)} \\
    \ell(L+1) &= -\frac{D}{2}\log(2\pi) - D\log\sigma(L+1) - \frac{D\,\varepsilon_{L+1}}{2\,\sigma^2(L+1)}
\end{align*}
Taking their difference and assuming the reconstruction error has saturated, i.e. $\varepsilon_L = \varepsilon_{L+1} = \varepsilon$, we have
\begin{align*}
\Delta\ell &= -D\log\frac{\sigma(L+1)}{\sigma(L)} + \frac{D\,\varepsilon_L}{2\,\sigma^2(L)} - \frac{D\,\varepsilon_{L+1}}{2\,\sigma^2(L+1)} \\
&= -D\log\gamma - \frac{D\,\varepsilon}{2\,\sigma^2(L)}\!\left(\frac{1}{\gamma^2} - 1\right)
\end{align*}
A crossover happens when the delta action log-likelihood $\Delta \ell = 0$. Solving for the corresponding $\varepsilon^*_L$:
\begin{align*}
    -D\log\gamma &= \frac{D\,\varepsilon}{2\,\sigma^2(L)}\!\left(\frac{1}{\gamma^2} - 1\right) \\
    \varepsilon^*_L &= \frac{-2\,\sigma^2(L)\,\log\gamma}{1/\gamma^2 - 1}
\end{align*}
For small $|\log\gamma|$, we have $1/\gamma^2 - 1 = e^{-2\log\gamma} - 1 \approx -2\log\gamma$. Therefore:
$$\varepsilon^*_L \approx \frac{-2\,\sigma^2(L)\,\log\gamma}{-2\log\gamma} = \sigma^2(L)$$
This suggests that the autoregressive process stops when the minimum achievable mean squared error at step $L$ is on the order of the decoder variance $\sigma^2(L)$. 

\paragraph{The regularization effect of the KL term} Without the KL term, the reconstruction objective alone admits a degenerate solution: the posterior could ``splat'' a multimodal action distribution with many small Gaussians, pushing the reconstruction error arbitrarily low and maxing out the latent steps. The KL term counteracts this degeneracy. While the posterior sees the input action and can route each sample to a dedicated latent path, the prior sees only the observation and must spread its probability mass to cover all paths. Under the splatting solution, the prior mass is stretched thin, and each routed sample incurs a KL cost that offsets the log-likelihood gain of the sharper Gaussians. This pulls the posterior back towards covering the action distribution with fewer, broader components, i.e., shorter latent traces. The characterization in Sec.~\ref{sec:compression} can therefore be stated more precisely: the model spends less compute when the remaining uncertainty cannot be reduced without a commensurate KL cost. We retain the simpler statement in the main text for clarity.
\section{Implementation Details}
\label{app:impl}
\subsection{Model Architecture}

Our implementation of \algnamepi{} consists of a shared latent encoder parameterizing the posterior $q_\theta(z\mid o, a)$ and the prior $p_\theta(z\mid o)$, and an action decoder parameterizing $p(a\mid o, z)$. $o$ is a stack of $h_o$ frames from $n_c$ camera views, each encoded via an image encoder into an $\ndim$-dimensional token. When provided, low-dimensional observations and language instructions are encoded using separate encoders into $\ndim$-dimensional tokens and concatenated with image tokens. $a$ is a chunk of $h_a$ actions, each encoded via a 2-layer MLP into an $\ndim$-dimensional action token. For the posterior $q_\theta(z\mid o, a)$, we concatenate observation and action tokens along the sequence dimension and pass them as cross-attention context to the causal transformer. For the prior $p_\theta(z\mid o)$, we mask out the action tokens with learned mask tokens. The latent encoder learns an $\nvocab + 1$ size codebook where the additional token denotes $\eos$. The autoregressive generation begins with a learnable root token $\texttt{ROOT}$ and iteratively generates the next token until $\eos$ is generated or a maximum sequence length $T$ is reached. The output sequence is then concatenated again with the observation embeddings and passed into the action decoder as cross-attention context. The action decoder is a bidirectional transformer with $h_a$ learnable query tokens. The corresponding output tokens are then decoded into action chunks via a 2-layer MLP. We use learned positional embeddings for both the encoder and the decoder. 

The action tokenizer \algnametok{} uses the same parameterization but drops the observation conditioning from both the encoder and the decoder. The downstream policy for the action tokenizer is an autoregressive transformer with an \eos{} token for early termination. Following~\cite{vqbet}, we train the autoregressive policy with a focal loss.

Detailed hyperparameters for \algnamepi{}, \algnametok{}, and autoregressive policy are presented in Table \ref{tab:hyperparams}.

\begin{table}[t]
    \centering
    \caption{Hyperparameters}
    \begin{minipage}[t]{0.49\linewidth}
        \centering
        \small
        \begin{tabular}{ll}
            \toprule
            \textbf{Model Parameter} & \textbf{Value} \\
            \midrule
            \textbf{\algnamepi{} Shared} &  \\
            Observation History $h_o$ & 1 \\
            Action Chunk $h_a$ & 16 \\
            Latent Sequence Length $H$ & 16 \\
            Num Registers & 1 \\
            MLP Ratio & 4 \\
            \midrule
            \textbf{\algnamepi{} Multi-Task} &  \\
            Latent Vocab Size $\nvocab$ & DROID: 64 \\
            & LIBERO: 16 \\
            Observation Encoder & SigLip (from Pi-0.5)\\
            Language Encoder & SigLip 2 \\
            Embed Dim $\ndim$ & 768 \\
            Encoder Depth & 12 \\
            Decoder Depth & 10 \\
            Num Heads & 12 \\
            Variance Range & DROID: 0.2-0.02 \\& LIBERO: 0.1-0.01\\
            \midrule
            \textbf{\algnamepi{} Single-Task} &  \\
            Latent Vocab Size $\nvocab$ & 16 \\
            Observation Encoder & ResNet-18 \\
            Embed Dim $\ndim$ & 384 \\
            Encoder Depth & 8 \\
            Decoder Depth & 6 \\
            Num Heads & 8 \\
            Variance Range & 0.1-0.01 \\
            \midrule
            \textbf{\algnametok} &  \\
            Latent Sequence Length $H$ & 8 \\
            Latent Vocab Size $\nvocab$ & 64 \\
            Initial Variance & 1.0  \\
            \midrule
            \textbf{Autoregressive Policy} &  \\
            Observation Encoder & ResNet-18 \\
            Embed Dim $\ndim$ & 1024 \\
            Depth & 12 \\
            Num Heads & 8 \\
            MLP Ratio & 4 \\
            \bottomrule
        \end{tabular}
    \end{minipage}
    \hfill
    \begin{minipage}[t]{0.49\linewidth}
        \centering
        \small
        \begin{tabular}{ll}
            \toprule
            \textbf{Training Parameter} & \textbf{Value ($\times$ GPUs)} \\
            \midrule
            \textbf{\algname{} Shared} \\
            Training Epochs & 2 \\
            Clip Range & 0.01 \\
            Prior Mixture Coef & 0.1 \\
            \midrule
            \textbf{\algnamepi{} Multi-Task} \\
            Num Samples & 5000000 $\times$ 32 \\
            Rollout Buffer Size & 3200 $\times$ 32 \\
            Batch Size & 64 $\times$ 32 \\
            Free Nats Ratio & 0.05 \\
            KL Loss Coef & 1.0 \\
            Reconstruction Loss Coef & 0.01 \\
            \midrule
            \textbf{\algnamepi{} Single-Task} &  \\
            Num Samples & 25000000 $\times$ 4 \\
            Rollout Buffer Size & 25000 $\times$ 4 \\
            Batch Size & 500 $\times$ 4 \\
            Free Nats Ratio & 0.05 \\
            KL Loss Coef & 1.0 \\
            Reconstruction Loss Coef & 0.1 \\
            \midrule
            \textbf{\algnametok} \\
            Free Nats Ratio & 0.2 \\
            KL Loss Coef & 0.1 \\
            Reconstruction Loss Coef & 1.0 \\
            \midrule
            \textbf{Autoregressive Policy} &  \\
            Number of gradient steps & 50000 $\times$ 4 \\
            Batch Size & 1024 $\times$ 4 \\
            Focal loss $\gamma$ & 1.0 \\
            \midrule
            \textbf{Optimization} &  \\
            Optimizer & AdamW \\
            Learning Rate & $1e^{-4}$ \\
            Weight Decay & $1e^{-6}$ \\
            Betas & [0.9, 0.999] \\
            Epsilon & $1e^{-8}$ \\
            \bottomrule
        \end{tabular}
    \end{minipage}
    \label{tab:hyperparams}
\end{table}

\subsection{Baseline Details}
In this section, we describe the details of baseline methods. All policy baselines use an observation history of 1 and action chunk of 16. We train all baselines for the same number of gradient steps as our method.

\paragraph{Diffusion Policy~\cite{diffusionpolicy} / Large Behavior Model~\cite{lbmtri2025}} are generative control policies which model the expert's action distribution using a denoising diffusion process. Diffusion Policy is the single-task variant, and Large Behavior Model is the language-conditioned multi-task variant. We parameterize the noise prediction network as a bidirectional transformer with cross-attention observation conditioning. We use 100 diffusion steps during training and 10 diffusion steps with DDIM scheduler for inference. 

\paragraph{VQ-BeT~\cite{vqbet}} is a generative control policy consisting of a residual VQ action tokenizer and an autoregressive transformer policy. We use the 2-layer residual VQ tokenizer from the official implementation of~\cite{vqbet} with a codebook size of 32 per layer. We then implement a 10-layer autoregressive transformer policy to regress the action tokens with cross-attention observation conditioning. Following~\cite{vqbet}, we train an additional residual policy conditioned on the continuous output embedding (used to decode the first-layer action token) to correct errors from the quantized actions.

\paragraph{VAE Policy} is a conditional Gaussian VAE~\cite{kingma2013vae} used as a non-iterative generative control policy baseline. The encoder and decoder are parameterized by bidirectional transformers with 8 and 6 layers respectively, and the observation conditioning is passed in as cross-attention context. The encoder predicts a continuous mean and log standard deviation of a Gaussian distribution. The action decoder outputs the means and a learned scalar variance as in~\cite{rybkin2021simple}. The model is trained end-to-end via reparameterization to optimize the ELBO. Instead of setting a fixed KL coefficient, we use a free-nats approach to constrain the KL within a set threshold of 0.1. 

\paragraph{FAST~\cite{pertsch2025fast}} is a variable-length discrete action tokenizer. The tokenization process (1) performs a Discrete Cosine Transform to map an action chunk to a frequency domain matrix, (2) quantizes the matrix into integers, (3) flattens the matrix into a sequence, and (4) applies byte pair encoding to chunk the sequence into tokens. We use the official implementation of~\cite{pertsch2025fast} with vocab size 1024 for the byte-pair encoding.

\paragraph{OAT~\cite{liu2026orderedactiontokenization}} is a variable-length latent action tokenizer which encodes continuous action chunks autoregressively into sequences of discrete tokens. The method injects a coarse-to-fine inductive bias by randomly dropping out suffix tokens and forcing early tokens to reconstruct the full actions. We use the official implementation of~\cite{liu2026orderedactiontokenization} with vocab size 1000 and sequence length 8.

\subsection{Environment Details}
\label{app:envs}
\begin{figure}
    \centering
    \includegraphics[width=\linewidth]{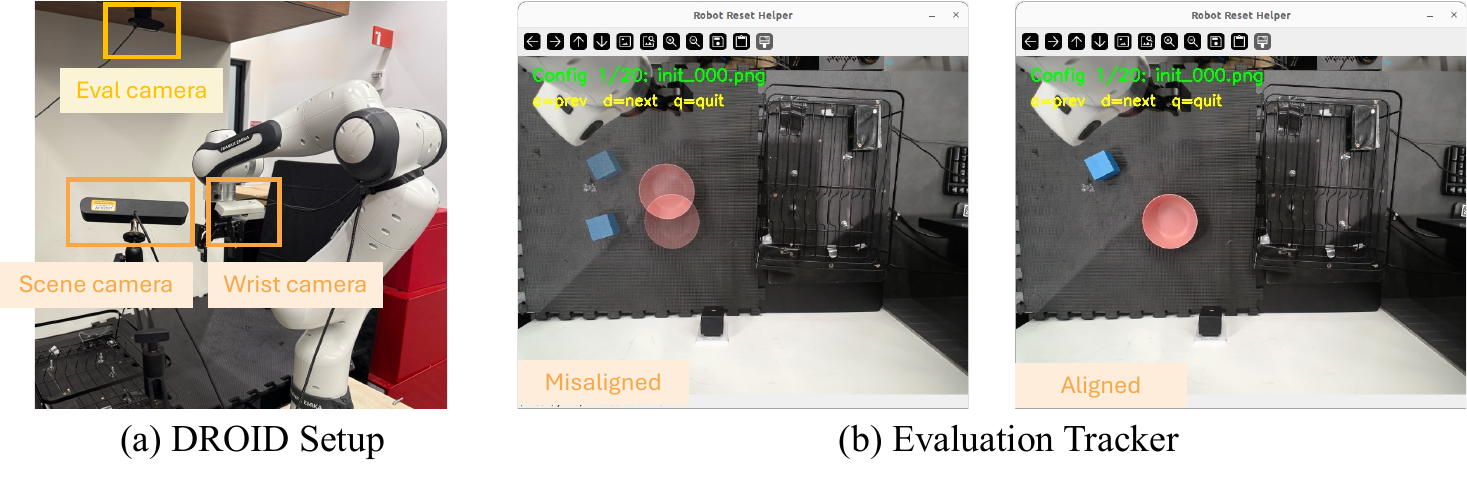}
    \caption{Real-world experiment setup. We run real-world experiments on the DROID Franka platform, with one external camera and one wrist camera. We mount an overhead camera to track the initializations for a controlled comparison between methods.}
    \label{fig:robot-setup}
\end{figure}

\paragraph{DROID~\cite{droid}} (Fig. \ref{fig:robot-setup}) is a single-arm manipulation platform based on a 7-DoF Franka robot. The image observations consist of two (180, 320) images from a scene camera and a wrist camera. The low-dimensional observations consist of 6-dimensional absolute joint positions and a 1-dimensional absolute gripper position, represented as a continuous value in [0, 1]. The actions consist of 6-dimensional delta joint positions and a 1-dimensional absolute gripper position. The controller runs at 10 Hz. To ensure reliable comparisons, we mount an overhead camera and use a program to track 20 initial configurations for each task. The program alpha-blends the pre-captured snapshot and the live frame so that objects can be reset by aligning them.
We evaluate on two zero-shot tasks roughly within the training distribution of the DROID dataset, and two challenging tasks where we collect 100 demonstrations each and finetune the policies.
\begin{enumerate}
    \item \textbf{block-bowl (zero-shot)} involves picking up a cube block and placing it in a bowl. The objects are randomly initialized across the workspace. An episode is considered successful if the gripper releases the block into the bowl. Language instruction: ``Pick up the block and put it in the bowl.''
    \item \textbf{marker-mug (zero-shot)} involves picking up a marker and dropping it in a mug. The objects are randomly initialized across the workspace. An episode is considered successful if the gripper releases the marker into the mug. Language instruction: ``Pick up the marker and put it in the mug.''
    \item \textbf{peg-hole (finetuned)} is a high-precision manipulation task which involves inserting a square peg into a base with a square hole. The base is randomly initialized in a square region near the center of the workspace, and the peg is randomly initialized outside the square region. An episode is considered successful if the gripper inserts the peg securely into the base. Language instruction: ``Insert the peg into the hole.''
    \item \textbf{clean-table (finetuned)} is a compositional manipulation task which involves placing 3 objects scattered around the workspace into a bowl. We collect demonstrations by randomly choosing 3 objects from a 7-object suite. We measure partial success of cleaning up at least 1, 2, and 3 objects. Language instruction: ``Clean up the table.''
\end{enumerate}

\paragraph{LIBERO~\cite{liu2023libero}} is a simulated robotic benchmark designed to evaluate lifelong learning algorithms. The benchmark involves controlling a 7-DoF Franka Panda robot to complete various tasks across different scenes. The LIBERO-100 suite consists of 100 tasks distributed across three scenes (kitchen, living room, study), split into 90 training tasks (LIBERO-90) and 10 held-out tasks (LIBERO-10). The observations are (128, 128) images from a scene camera and a wrist camera, and the actions are 7-dimensional delta endeffector poses.

\paragraph{D3IL~\cite{jia2024d3il}} is a simulated robotic benchmark designed to evaluate multimodal behaviors. The benchmark features a Franka Panda robot with task-specific endeffectors. We use tasks that provide vision data for our experiments. The observations are (96, 96) images from a scene camera and a wrist camera. The actions are absolute endeffector poses whose dimensionality varies across tasks. We describe the individual tasks below. 

\begin{enumerate}
    \item \textbf{avoiding} is an analytical task where the robot needs to navigate from one end of the table to the other while avoiding the rods arranged in a pyramidal shape. The gaps between the rods naturally create multimodality. The action space is 2-dimensional xy coordinates.
    \item \textbf{aligning} involves aligning a hollow square shape to a target pose with a rod endeffector, where colored faces need to match. The action space is 3-dimensional xyz coordinates. The strategy is to either push from outside the square, or insert the rod inside the square and pull. The dataset contains both pushing and pulling demonstrations. 
    \item \textbf{stacking} involves stacking three cubes (red, blue, green) in a target region using a parallel-jaw gripper. The action space is 8-dimensional endeffector pose. The dataset contains demonstrations of stacking the objects in different orders. 
    \item \textbf{sorting (2, 4, 6)} involves pushing red and blue colored blocks into corresponding red and blue colored bins using a rod endeffector. The 2, 4, and 6 variants have 1, 2, and 3 blocks for each color. The action space is 2-dimensional xy coordinates. The dataset contains demonstrations of sorting the objects in different orders. 
\end{enumerate}

\paragraph{Robomimic~\cite{mandlekar2021robomimic}} is a simulated robotic benchmark which involves controlling a Franka Panda arm to complete tasks of varying difficulty. We use the ph dataset for all our experiments. The actions are 7-dimensional delta endeffector poses for all single-arm tasks and 14-dimensional for bimanual tasks. The observations vary across tasks. We describe the individual tasks below. 

\begin{enumerate}
    \item \textbf{lift} involves lifting a cube off the table. The observations are (84, 84) images from a scene camera and a wrist camera.
    \item \textbf{can} involves picking up a can from one table and putting it in the correct cell in the adjacent table. The observations are (84, 84) images from a scene camera and a wrist camera.
    \item \textbf{square} involves picking up a ``frame'' piece with a square hole and dropping it on a square peg. The observations are (84, 84) images from a scene camera and a wrist camera.
    \item \textbf{tool-hang} is a high-precision task that involves inserting a hanger on a narrow stand and hanging a wrench on the hanger. The observations are (240, 240) images from a scene camera and a wrist camera. 
    \item \textbf{transport} is a bimanual task involving a sequence of operations: the left arm lifts up the lid and hands the hammer to the right arm, while the right arm first moves a block from one bin to another, then receives the hammer and places it into a bin. The observations are (84, 84) images from one scene camera and two wrist cameras. 
\end{enumerate}
\section{Additional Experiments}
\label{app:exps}

\subsection{Distribution of task success rates across LIBERO-90}
\begin{figure}[t]
    \centering
    \includegraphics[width=\linewidth]{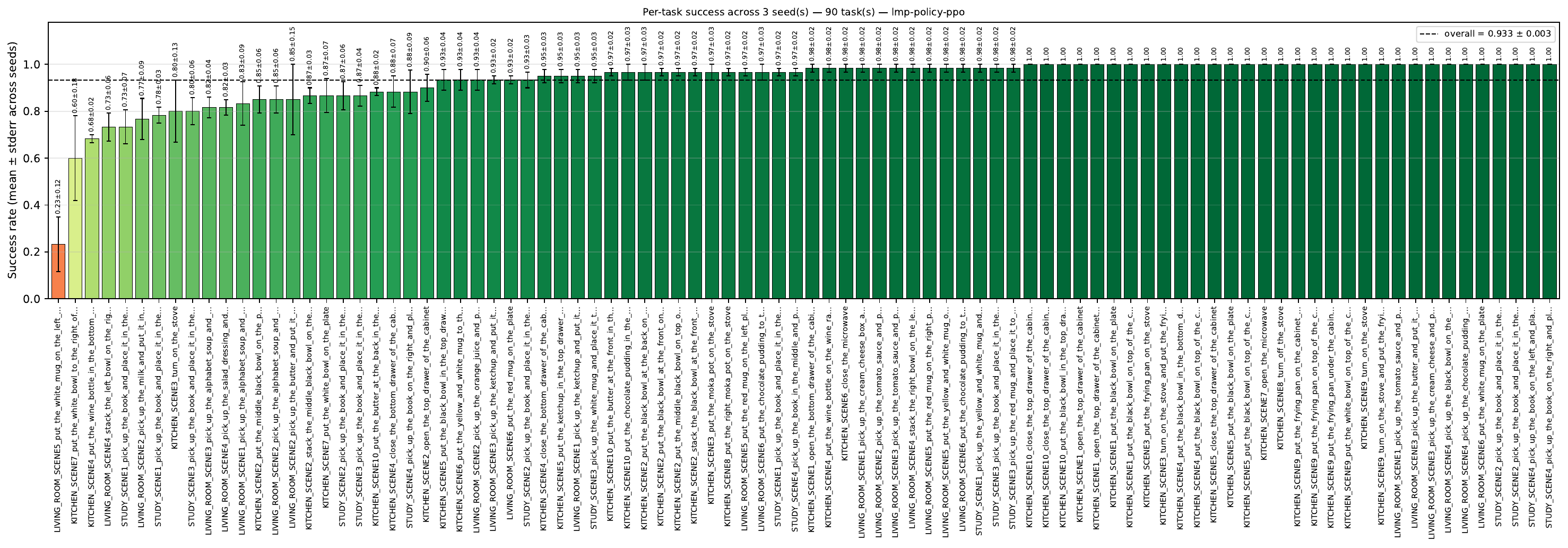}
    \includegraphics[width=\linewidth]{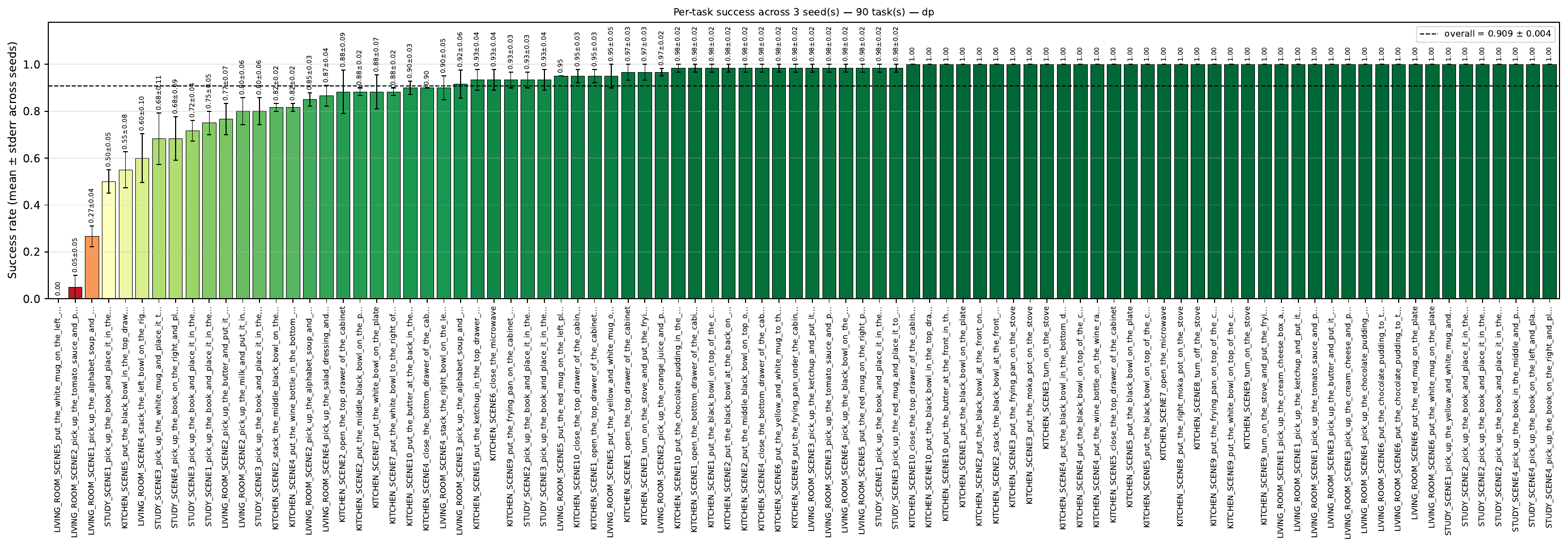}
    \caption{Distribution of single-task success rates over LIBERO-90. \algnamepi{} (top) shows improved generalization across tasks compared to DP (bottom).}
    \label{fig:per-task-success}
\end{figure}

In Fig. \ref{fig:per-task-success}, we plot the per-task success rates over all 90 tasks in LIBERO-90 over 3 seeds. While both \algnamepi{} and DP achieve near-perfect success in a majority of tasks, \algnamepi{} tapers off at a slower rate, marking better generalization across task semantics and less cross-task interference.

\subsection{Full single-task results}

\begin{table*}[t]
\centering
\small
\caption{Evaluation Results Across Robomimic and D3IL Tasks}
\begin{tabular}{lcccc}
\toprule
Task & VQ-BET & VAE & DP & \algnamepi{} (Ours) \\
\midrule
d3il-aligning       & 0.23 \std{0.03}   & 0.51 \std{0.07}   & \textbf{0.74 \std{0.04}}   & \textbf{0.71 \std{0.04}}   \\
d3il-stacking       & 0.53 \std{0.03}   & 0.71 \std{0.07}   & 0.81 \std{0.01}   & \textbf{0.86 \std{0.04}}   \\
d3il-sorting-2      & 0.46 \std{0.02}   & 0.67 \std{0.13}   & 0.75 \std{0.01}   & \textbf{0.80 \std{0.01}}   \\
d3il-sorting-4      & 0.41 \std{0.01}   & 0.68 \std{0.04}   & \textbf{0.69 \std{0.05}}   & \textbf{0.68 \std{0.02}}   \\
d3il-sorting-6      & 0.32 \std{0.04}   & 0.70 \std{0.04}   & 0.68 \std{0.03}   & \textbf{0.75 \std{0.04}}   \\
\midrule
d3il average        & 0.39 & 0.57 & 0.73 & \textbf{0.76} \\
\midrule
robomimic-lift      & 0.96 \std{0.00}   & 0.97 \std{0.01}   & \textbf{1.00 \std{0.00}}   & \textbf{1.00 \std{0.00}}   \\
robomimic-can       & 0.83 \std{0.04}   & 0.78 \std{0.07}   & \textbf{1.00 \std{0.00}}   & \textbf{1.00 \std{0.00}}   \\
robomimic-square    & 0.58 \std{0.06}   & 0.79 \std{0.08}   & \textbf{0.92 \std{0.00}}   & 0.87 \std{0.02}   \\
robomimic-tool-hang & 0.15 \std{0.04}   & 0.29 \std{0.06}   & \textbf{0.53 \std{0.07}}   & \textbf{0.51 \std{0.08}}   \\
robomimic-transport & 0.26 \std{0.05}   & 0.58 \std{0.12}   & \textbf{0.80 \std{0.10}}   & \textbf{0.75 \std{0.08}}   \\
\midrule
robomimic average   & 0.56 & 0.68 & \textbf{0.85} & 0.83 \\
\bottomrule
\end{tabular}
\label{tab:exp-policy-singletask}
\end{table*}

Table \ref{tab:exp-policy-singletask} presents the full numerical results on single-task benchmarks from Sec. \ref{exp:non-iterative}. As discussed in Sec. \ref{exp:non-iterative}, non-iterative baselines using a similar latent-space formulation fail to capture the granularity required to solve complex manipulation tasks. We additionally compare to Diffusion Policy~\cite{diffusionpolicy}, an iterative baseline which models the action generation process with a denoising diffusion process. We find DP to achieve comparable performance to \algnamepi{} in single-task domains.

\subsection{Full tokenizer results}
\begin{table}[t]
\centering
\small
\caption{Tokenizer Evaluation Results Across Robomimic and LIBERO Tasks.}
\begin{tabular}{lcccc}
\toprule
Task & VQVAE & FAST & OAT & \algnametok{} \\
\midrule
robomimic-lift      & 0.82 \std{0.06}   & 0.98 \std{0.02}   & 0.97 \std{0.02}  & \textbf{1.00 \std{0.00}}   \\
robomimic-can       & 0.32 \std{0.08}   & 0.57 \std{0.12}   & 0.57 \std{0.08}  & \textbf{0.93 \std{0.02}}   \\
robomimic-square    & 0.02 \std{0.02}   & 0.42 \std{0.10}   & 0.32 \std{0.08}  & \textbf{0.67 \std{0.10}}   \\
robomimic-tool-hang & 0.00 \std{0.00}   & 0.00 \std{0.00}   & 0.00 \std{0.00}  & \textbf{0.17 \std{0.02}}   \\
\midrule
robomimic average   & 0.29              & 0.49              & 0.47              & \textbf{0.69}             \\
\midrule
libero-soup-cheese  & 0.18 \std{0.05}   & 0.22 \std{0.13}   & 0.15 \std{0.04}  & \textbf{0.92 \std{0.05}}   \\
libero-mug-mug      & 0.28 \std{0.02}   & 0.10 \std{0.04}   & 0.23 \std{0.02}  & \textbf{0.88 \std{0.06}}   \\
libero-moka-moka    & 0.00 \std{0.00}   & 0.02 \std{0.02}   & 0.17 \std{0.08}  & \textbf{0.63 \std{0.08}}   \\
libero-bowl-drawer  & 0.02 \std{0.02}   & 0.37 \std{0.09}   & 0.60 \std{0.07}  & \textbf{0.75 \std{0.08}}   \\
libero-book-caddy   & 0.23 \std{0.06}   & 0.40 \std{0.07}   & 0.42 \std{0.10}  & \textbf{0.57 \std{0.06}}   \\
\midrule
libero average      & 0.14              & 0.22              & 0.31              & \textbf{0.75}             \\
\bottomrule
\end{tabular}
\label{tab:exp-tokenizer}
\end{table}

Table \ref{tab:exp-tokenizer} shows the full numerical results of tokenizer experiments on Robomimic and LIBERO tasks over 3 seeds. As discussed in Sec. \ref{exp:tokenizer}, we pretrain the methods on a multi-task dataset and use them to train downstream autoregressive policies. For Robomimic, we train on the combined dataset of single-arm tasks: lift, can, square, tool-hang, and evaluate on the same tasks. For LIBERO, we train on the combined dataset of LIBERO-90 and evaluate on LIBERO-10. When controlling for the downstream policy, \algnametok{} achieves significantly higher success rates than baselines.

\subsection{Additional qualitative analysis of test-time compute} 
\label{app:exp-test-time-compute}

\begin{figure*}[t]
    \centering
    \includegraphics[width=\linewidth]{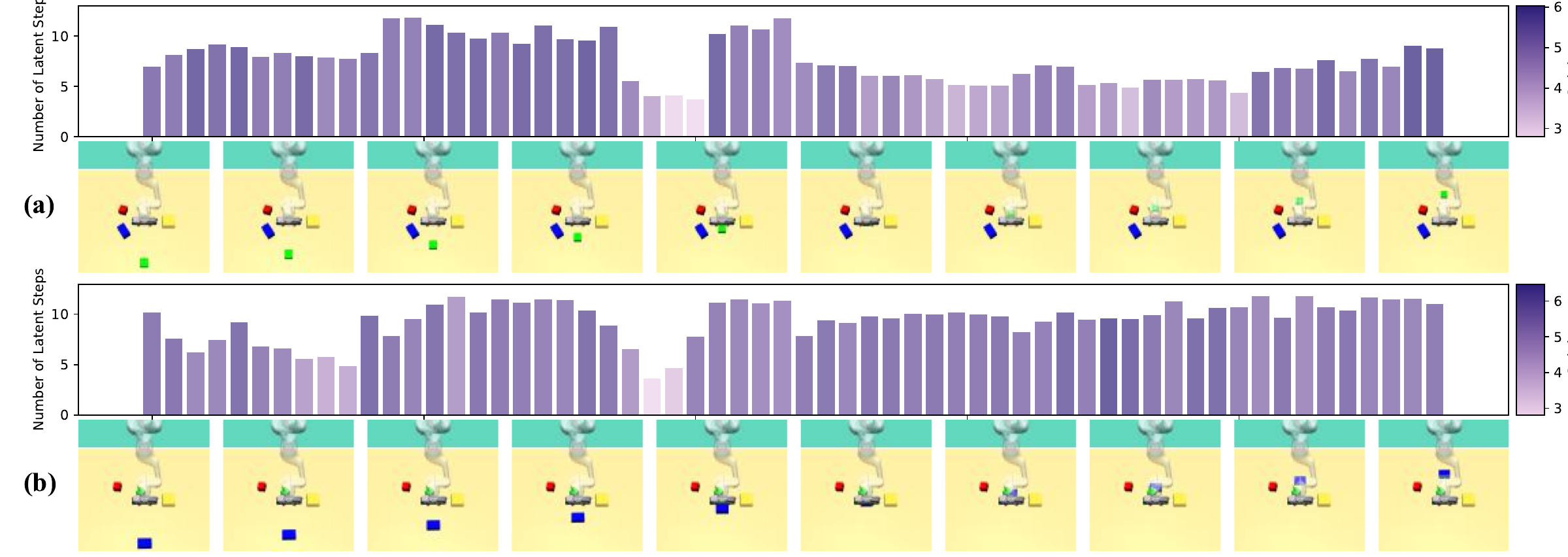}
    \caption{Analysis of test-time compute allocation. We manually move a block through an open gripper while querying the model at each observation. The number of latent steps reaches the lowest point as the block passes through the gripper. }
    \label{fig:analysis-scan}
\end{figure*}

In Fig. \ref{fig:analysis-scan}, we construct an analytical experiment in d3il-stacking to qualitatively understand the allocation of test-time compute. We keep the gripper in an open position and slide a block through the open gripper, while continuously querying the policy for the average number of latent steps. As the block moves, the number of latent steps reaches the lowest point the moment the block is right underneath the gripper. This qualitatively shows a correlation between reasoning steps and potential gripper movement.

\begin{figure*}
    \centering
    \includegraphics[width=\linewidth]{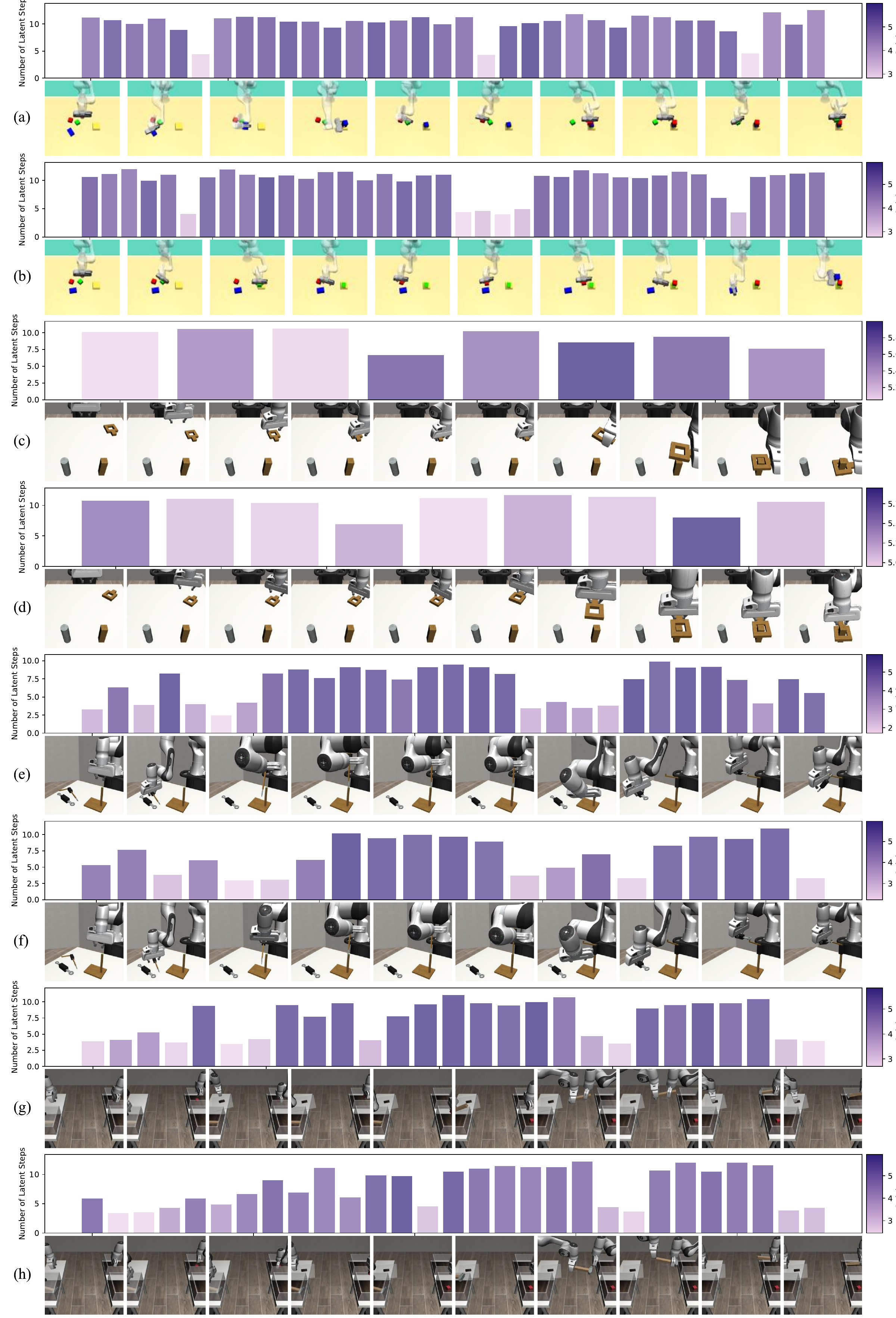}
    \caption{Additional visualization of test-time compute allocation in simulation tasks. Each timestep is averaged over 64 model queries. (a)-(b): d3il-stacking. (c)-(d): robomimic-square. (e)-(f) robomimic-tool-hang. (g)-(h): robomimic-transport.}
    \label{fig:app-vis}
\end{figure*}

\begin{figure*}
    \centering
    \includegraphics[width=\linewidth]{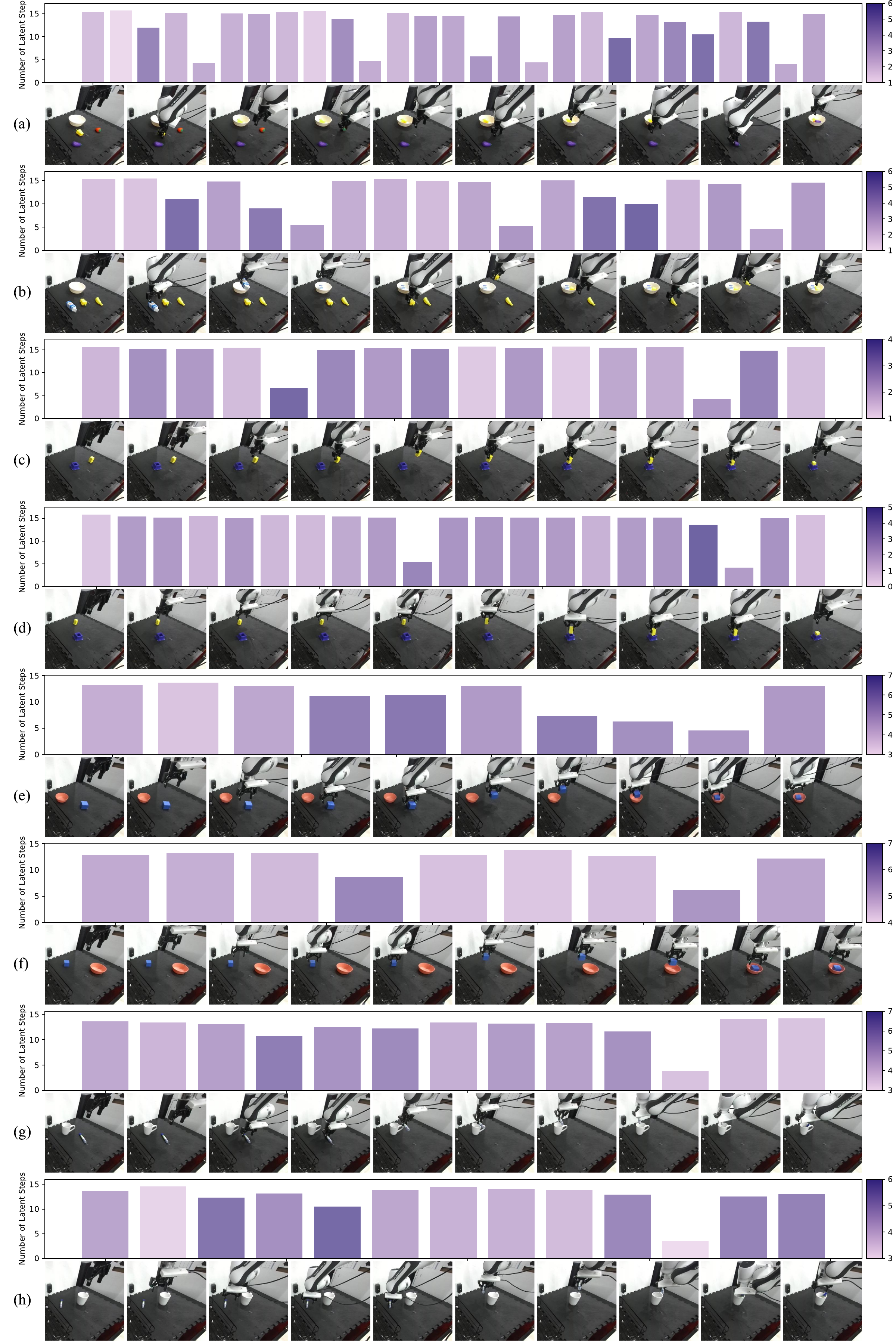}
    \caption{Additional visualization of test-time compute allocation in real robot tasks. Each timestep is averaged over 64 model queries. (a)-(b) clean-table (finetuned). (c)-(d): peg-hole (finetuned). (e)-(f) block-bowl (zero-shot). (g)-(h): marker-mug (zero-shot).}
    \label{fig:app-vis-droid}
\end{figure*}

We provide additional visualizations of episodic reasoning traces in Fig. \ref{fig:app-vis} and \ref{fig:app-vis-droid}. We find that across a number of simulated and real-world tasks, the number of latent steps is low during grasping and releasing, and high during aligning and other movements. We did not observe as much latent step variation in simpler tasks such as Robomimic lift and can, or D3IL aligning and sorting where the endeffector is a peg.

\subsection{Visualization of training dynamics}
\begin{wrapfigure}{r}{0.4\textwidth}
    \vspace{-0.3cm}
    \centering
    \includegraphics[width=\linewidth]{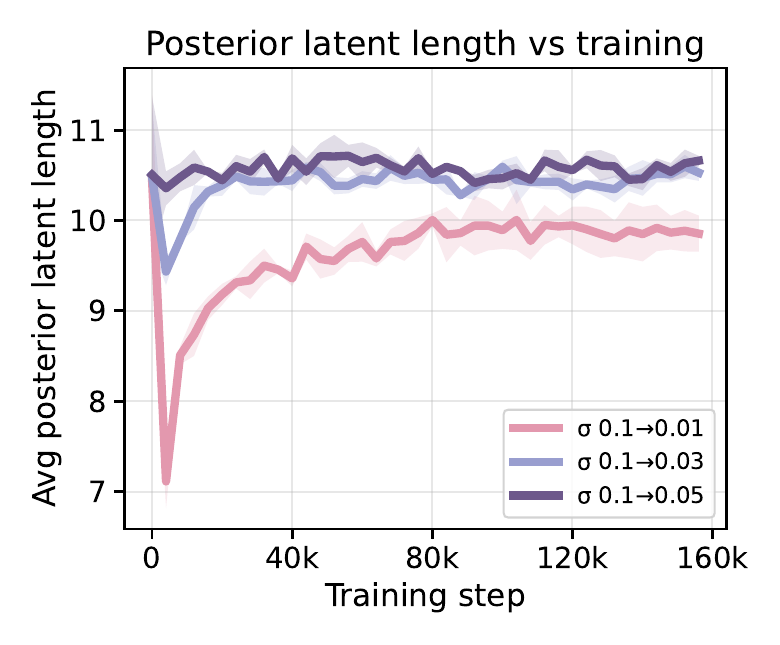}
    \caption{Average latent length vs training steps. As training progresses, the model demonstrates spontaneous increase in average latent steps.}
    \label{fig:analysis-training-latent-length}
\end{wrapfigure}

To understand the training dynamics of \algnamepi{}, we plot the average number of reasoning steps against the number of gradient steps in Fig. \ref{fig:analysis-training-latent-length}. At the beginning of training, the model is randomly initialized, resulting in a uniform latent distribution with a high average length. Early in training, the model is dominated by the reconstruction objective, collapsing to short reasoning traces to myopically maximize the action log likelihood. As training progresses, the length of latent trace gradually increases as the model learns to utilize more latent steps to effectively reconstruct the action distribution. Notably, the training curve resembles DeepSeek R1~\cite{deepseekr1}, where reasoning length increases over training. We also find that more aggressive decoder variance decay leads to shorter average reasoning lengths, confirming the effectiveness of compression as a length penalty.

\subsection{Ablation of latent steps}
\begin{wrapfigure}{r}{0.4\linewidth}
    \vspace{-0.5cm}
    \centering
    \includegraphics[width=\linewidth]{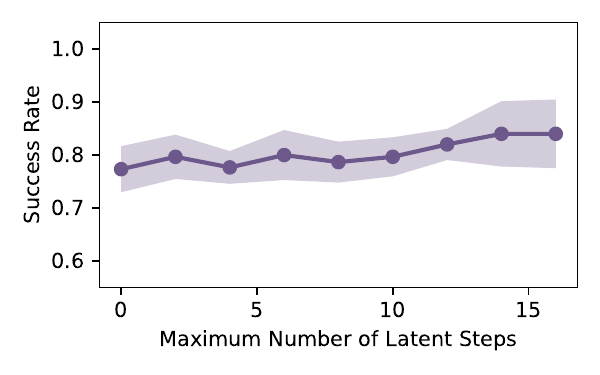}
    \caption{Success rate vs latent steps. Truncating the number of latent steps at test time leads to a marginal drop in performance. This happens when the model exhibits mode-seeking behavior.}
    \label{fig:ablation-latent-step}
    \vspace{-1.0cm}
\end{wrapfigure}
In Fig. \ref{fig:ablation-latent-step}, we plot the success rates of d3il-stacking policies as we decrease the maximum number of latent steps at test time. This is done by truncating the autoregressive sampling at the desired step and decoding the partial traces into actions. To our surprise, we find the performance to drop marginally as the number of latent steps decreases. We attribute this to the model exhibiting mode-seeking behavior: although the policy commits to fewer modes, each mode still decodes to a valid action. For environments with deterministic dynamics, a reduction in multimodality is not reflected in lower success rates. This is consistent with our finding in Fig. \ref{fig:ablation-multimodality}, where we visualize the mode-seeking behavior in d3il-avoiding.

\subsection{Ablation of latent dimension}
\begin{wraptable}{r}{0.4\linewidth}
\vspace{-0.5cm}
\centering
\caption{Ablation of Latent Dimension.}
\begin{tabular}{ccc}
\toprule
Vocab Size  & Seq Len & d3il-stacking \\
\midrule
2  &  4     &  0.60 \std{0.07} \\
2  &  8     &  0.71 \std{0.01}\\
2  &  16    &  0.72 \std{0.02}\\
\midrule
4  & 4      &  0.78 \std{0.05} \\
4  & 8      &  0.76 \std{0.02}\\
4  & 16     &  0.80 \std{0.01} \\
\midrule
16  & 16    &  0.86 \std{0.04} \\ 
\bottomrule
\end{tabular}
\label{tab:exp-ablate-latent-size}
\end{wraptable}

In Table \ref{tab:exp-ablate-latent-size}, we conduct an ablation experiment to analyze the impact of the latent space dimension on performance. We evaluate \algnamepi{} with different vocabulary sizes (width) and sequence lengths (depth) on d3il-stacking and report the average success rates. We find performance to drop as the vocab size and sequence length decrease, signaling insufficient representation capacity. On this particular task, we find performance to saturate with a latent space as small as vocab size 4 and sequence length 4. We use a larger latent space of sequence length 16 and vocab size 16 in our experiment to ensure that the latent capacity is not a bottleneck.

\subsection{Token allocation in autoregressive policies trained on \algnametok{}}

\begin{figure}
    \centering
    \includegraphics[width=\linewidth]{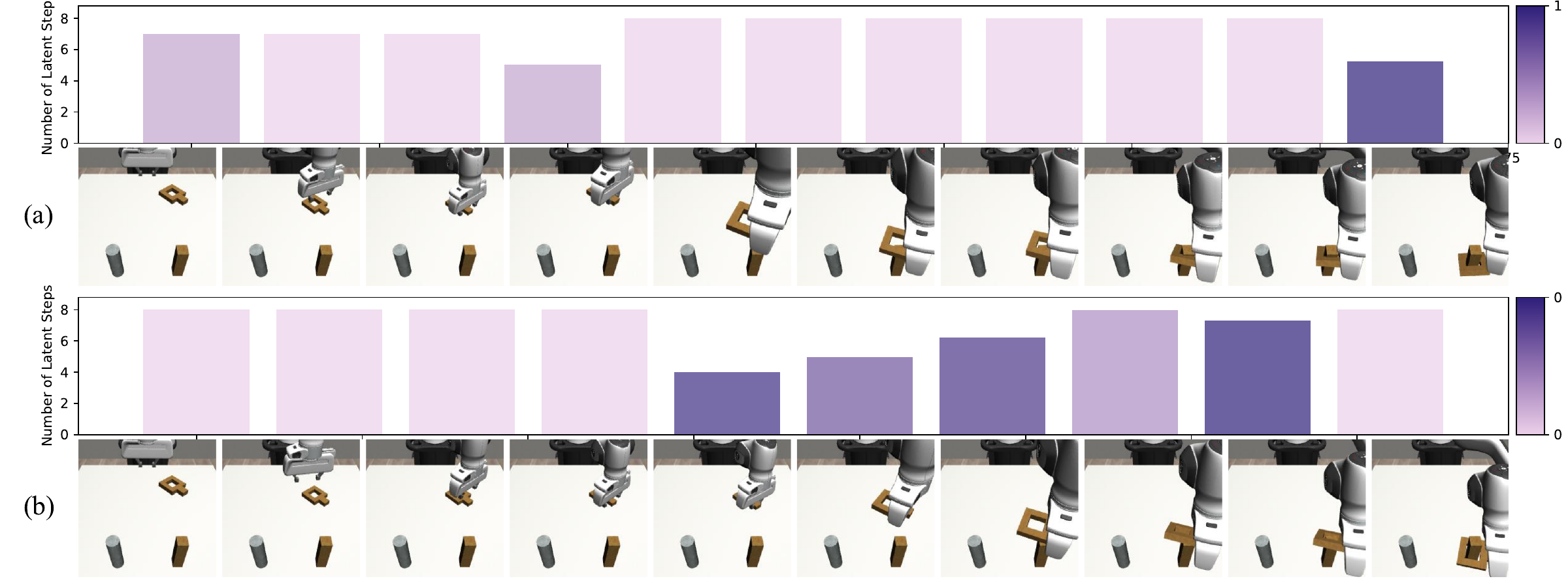}
    \caption{Allocation of action tokens in autoregressive policies trained on \algnametok{}. (a) When \algnametok{} is trained with a length penalty, the downstream autoregressive policy shows similar ``reasoning'' behavior to \algnamepi{} where gripper movements correspond to fewer action tokens. (b) Removing the length penalty results in more arbitrary allocation of action tokens.}
    \label{fig:tokenizer-allocation}
\end{figure}
In Fig. \ref{fig:tokenizer-allocation}, we visualize the allocation of action tokens in autoregressive policies trained on \algnametok{}. When \algnametok{} is trained with a length penalty ($\sigma$ decaying from 1.0-0.1), the downstream autoregressive policy behaves similarly to \algnamepi{}, where gripper movements are associated with fewer action tokens. However, the allocation of action tokens is rather arbitrary when the tokenizer is trained without a length penalty (learned scalar $\sigma$). Since we do not find the length penalty to improve policy performance, we use the simpler scalar variance for our experiments in Sec. \ref{exp:tokenizer}.

\section{Pseudocode}

We provide the pseudocode for \algnamepi{} in Alg. \ref{alg:lmp-pi} and \algnametok{} in Alg. \ref{alg:lmp-tok}.

\newcommand{\Dexp}{\mathcal D_\texttt{exp}}
\newcommand{\Dbuf}{\mathcal D_\texttt{buf}}
\newcommand{\Dact}{\mathcal D_\texttt{act}}

\begin{algorithm}
\caption{Latent Memory Palace Policy (\algnamepi{})}
\label{alg:lmp-pi}
    \begin{algorithmic}[1]
    \INPUT Expert dataset $\Dexp$
    \STATE Initialize neural network parameters $\theta, \phi$.
    \WHILE{not converged}
        \STATE \texttt{// Collect on-policy rollouts}
        \STATE Draw $|\Dbuf|$ expert samples $\{(o_i, a_i)\}_{i=1}^{|\Dbuf|} \sim \Dexp$.
        \FOR{$i = 1 \dots |\Dbuf|$}
        \STATE Sample latent traces from the posterior $z_i \sim q_\theta(\cdot|o_i, a_i)$
        \STATE Push $z_i$ and the on-policy log likelihood $\log q_\theta^{\texttt{old}}(z_i|o_i, a_i)$ to buffer $\Dbuf$.
        \ENDFOR
        \STATE \texttt{// Update model}
        \FOR{epoch $e = 1 \dots N$}
            \FOR{minibatch $\{(o_j, a_j, z_j, \log q_\theta^{\texttt{old}}(z_j|o_j, a_j))\}_{j=1}^{\mathcal B} \sim \Dbuf$}
                \STATE Forward pass through the latent encoder to compute $\log q^{\texttt{new}}_\theta(z_j | o_j, a_j)$ and $\log p_\theta(z_j | o_j)$ 
                \STATE Forward pass through the action decoder to obtain $\log p_\phi(a_j|o_j, z_j)$
                \STATE Compute objective $J(\theta, \phi)$ from Eq. \ref{eq:ppo_surrogate}
                \STATE Update $\theta$ and $\phi$ using gradient descent. 
            \ENDFOR
        \ENDFOR
    \ENDWHILE
    \end{algorithmic}
\end{algorithm}

\begin{algorithm}
\caption{Latent Memory Palace Action Tokenizer (\algnametok{})}
\label{alg:lmp-tok}
    \begin{algorithmic}[1]
    \INPUT Action dataset $\Dact$
    \STATE Initialize neural network parameters $\theta, \phi$.
    \WHILE{not converged}
        \STATE \texttt{// Collect on-policy rollouts}
        \STATE Draw $|\Dbuf|$ action samples $\{(a_i)\}_{i=1}^{|\Dbuf|} \sim \Dact$.
        \FOR{$i = 1 \dots |\Dbuf|$}
            \STATE Sample latent traces from the posterior $z_i \sim q_\theta(\cdot|a_i)$ 
            \STATE Push $z_i$ and the on-policy log likelihood $\log q_\theta^{\texttt{old}}(z_i|a_i)$ to buffer $\Dbuf$.
        \ENDFOR
        \STATE \texttt{// Update model}
        \FOR{epoch $e = 1 \dots N$}
            \FOR{minibatch $\{(o_j, a_j, z_j, \log q_\theta^{\texttt{old}}(z_j|a_j))\}_{j=1}^{\mathcal B} \sim \Dbuf$}
                \STATE Forward pass through the latent encoder to compute $\log q^{\texttt{new}}_\theta(z_j | a_j)$
                \STATE Forward pass through the action decoder to obtain $\log p_\phi(a_j|z_j)$
                \STATE Compute objective $J(\theta, \phi)$ from Eq. \ref{eq:ppo_surrogate_tokenizer}
                \STATE Update $\theta$ and $\phi$ using gradient descent. 
            \ENDFOR
        \ENDFOR
    \ENDWHILE
    \end{algorithmic}
\end{algorithm}

\end{document}